\documentclass{article} 
\usepackage{iclr2023_conference,times}

\usepackage{amsmath,amsfonts,bm}

\def\eqref#1{equation~\ref{#1}}

\def\1{\bm{1}}

\DeclareMathAlphabet{\mathsfit}{\encodingdefault}{\sfdefault}{m}{sl}
\SetMathAlphabet{\mathsfit}{bold}{\encodingdefault}{\sfdefault}{bx}{n}

\newcommand{\E}{\mathbb{E}}

\newcommand{\Cov}{\mathrm{Cov}}

\usepackage{hyperref}
\usepackage{url}
\usepackage{microtype}

\usepackage{graphicx}
\usepackage{subfigure}
\usepackage{booktabs} 
\usepackage{authblk}

\usepackage{multirow}
\usepackage{multicol}
\usepackage{algorithm}
\usepackage{algpseudocode}
\usepackage[titletoc,toc,title]{appendix}

\usepackage{amsmath}
\usepackage{amssymb}
\usepackage{xcolor}
\usepackage{dirtytalk}
\usepackage{color, soul}

\usepackage[utf8]{inputenc} 
\usepackage[T1]{fontenc}    
\usepackage{hyperref}       
\usepackage{url}            
\usepackage{booktabs}       
\usepackage{amsfonts}       
\usepackage{nicefrac}       
\usepackage{microtype}      
\usepackage{xcolor}         

\definecolor{nips-update}{RGB}{0,0,255}

\newtheorem{theorem}{Theorem}
\newtheorem{lemma}{Lemma}

\newtheorem{assumption}{Assumption}

\newcommand{\Prob}{\text{Pr}}

\title{\centering Towards scalable and non-IID robust \\
Hierarchical Federated Learning
via Label-driven Knowledge Aggregator}

\author[1]{Minh-Duong~Nguyen}
\author[2]{Q.-Viet~Pham}
\author[3]{Dinh Thai Hoang}
\author[4]{Long Tran-Thanh}
\author[3]{Diep~N.~Nguyen}
\author[1]{Won-Joo Hwang \thanks{Corresponding author.}}

\affil[1]{Pusan National University}
\affil[2]{Pusan National University}
\affil[4]{University of Warwick}
\affil[3]{University of Technology Sydney}

\newcommand{\fix}{\marginpar{FIX}}
\newcommand{\new}{\marginpar{NEW}}

\iclrfinalcopy

\begin{document}

\maketitle

\begin{abstract}
In real-world applications, Federated Learning (FL) meets two challenges: (1) scalability, especially when applied to massive IoT networks, and (2) how to be robust against an environment with heterogeneous data. Realizing the first problem, we aim to design a novel FL framework named Full-stack FL (F2L). More specifically, F2L utilizes a hierarchical network architecture, making extending the FL network accessible without reconstructing the whole network system. Moreover, leveraging the advantages of hierarchical network design, we propose a new label-driven knowledge distillation (LKD) technique at the global server to address the second problem. As opposed to current knowledge distillation techniques, LKD is capable of training a student model, which consists of good knowledge from all teachers' models. Therefore, our proposed algorithm can effectively extract the knowledge of the regions' data distribution (i.e., the regional aggregated models) to reduce the divergence between clients' models when operating under the FL system with non-independent identically distributed data.
Extensive experiment results reveal that: (i) our F2L method can significantly improve the overall FL efficiency in all global distillations, and (ii) F2L rapidly achieves convergence as global distillation stages occur instead of increasing on each communication cycle.
\end{abstract}

\section{Introduction}
\label{Introduction}
Recently, Federated Learning (FL) is known as a novel distributed learning methodology for enhancing communication efficiency and ensuring privacy in traditional centralized one \cite{2017-FL-FederatedLearning}. However, the most challenge of this method for client models is non-independent and identically distributed (non-IID) data, which leads to divergence into unknown directions. Inspired by this, various works on handling non-IID were proposed in 
\cite{2020-FL-FedProx,2021-FL-FedDyne,2021-FL-FedU,2020-FL-Scaffold,2021-FL-FedNova,2021-FL-FedGen,2022-FL-Wasserstein}. However, these works mainly rely on arbitrary configurations without thoroughly understanding the models' behaviors, yielding low-efficiency results. Aiming to fulfil this gap, in this work, we propose a new hierarchical FL framework using information theory by taking a deeper observation of the model's behaviors, and this framework can be realized for various FL systems with heterogeneous data. In addition, our proposed framework can trigger the FL system to be more scalable, controllable, and accessible through hierarchical architecture. Historically, anytime a new segment (i.e., a new group of clients) is integrated into the FL network, the entire network must be retrained from the beginning. Nevertheless, with the assistance of LKD, the knowledge is progressively transferred during the training process without information loss owing to the empirical gradients towards the newly participated clients' dataset.

The main contributions of the paper are summarized as follows. \textbf{(1)} We show that conventional FLs performance is unstable in heterogeneous environments due to non-IID and unbalanced data by carefully analyzing the basics of Stochastic Gradient Descent (SGD). \textbf{(2)} We propose a new multi-teacher distillation model, Label-Driven Knowledge Distillation (LKD), where teachers can only share the most certain of their knowledge. In this way, the student model can absorb the most meaningful information from each teacher. \textbf{(3)} To trigger the scalability and robustness against non-IID data in FL, we propose a new hierarchical FL framework, subbed Full-stack Federated Learning (F2L). Moreover, to guarantee the computation cost at the global server, F2L architecture integrates both techniques: LKD and FedAvg aggregators at the global server. To this end, our framework can do robust training by LKD when the FL process is divergent (i.e., at the start of the training process). When the training starts to achieve stable convergence, FedAvg is utilized to reduce the server's computational cost while retaining the FL performance. \textbf{(4)} We theoretically investigate our LKD technique to make a brief comparison in terms of performance with the conventional Multi-teacher knowledge distillation (MTKD), and in-theory show that our new technique always achieves better performance than MTKD. \textbf{(5)} We validate the practicability of the proposed LKD and F2L via various experiments based on different datasets and network settings. To show the efficiency of F2L in dealing with non-IID and unbalanced data, we provide a performance comparison and the results show that the proposed F2L architecture outperforms the existing FL methods. Especially, our approach achieves comparable accuracy when compared with FedAvg (\cite{2017-FL-FederatedLearning}) and higher $7-20\%$ in non-IID settings. 

\section{Related Work}
\label{sec:related-works}
\subsection{Federated Learning on non-IID data}
\label{sec:fl-non-iid}
To narrow the effects of divergence weights, some recent studies focused on gradient regularization aspects \cite{2020-FL-FedProx,2021-FL-FedDyne,2021-FL-FedU,2020-FL-Scaffold,2021-FL-FedNova,2021-FL-FedGen,2022-FL-Wasserstein}. By using the same conceptual regularization, the authors in \cite{2020-FL-FedProx,2021-FL-FedDyne}, and \cite{2021-FL-FedU} introduced the FedProx, FedDyne, and FedU, respectively, where FedProx and FedDyne focused on pulling clients' models back to the nearest aggregation model while FedU's attempted to pull distributed clients together. To direct the updated routing of the client model close to the ideal server route, the authors in \cite{2020-FL-Scaffold} proposed {SCAFFOLD} by adding a control variate to the model updates. Meanwhile, to prevent the aggregated model from following highly biased models, the authors in \cite{2021-FL-FedNova} rolled out FedNova by adding gradient scaling terms to the model update function. Similar to \cite{2021-FL-FedU}, the authors in \cite{2022-FL-Wasserstein} launched the WALF by applying Wasserstein metric to reduce the distances between local and global data distributions. However, all these methods are limited in providing technical characteristics. For example, \cite{2021-FL-FedNova} demonstrated that FedProx and FedDyne are ineffective in many cases when using pullback to the globally aggregated model. Meanwhile, FedU and WAFL have the same limitation on making a huge communication burden. Aside from that, FedU also faces a very complex and non-convex optimization problem.

Regarding the aspect of knowledge distillation for FL, only the work in \cite{2021-FL-FedGen} proposed a new generative model of local users as an alternative data augmentation technique for FL. However, the majority drawback of this model is that the training process at the server demands a huge data collection from all users, leading to ineffective communication.

Motivated by this, we propose a new FL architecture that is expected to be more elegant, easier to implement, and much more straightforward. Unlike \cite{2021-FL-FedU, 2021-FL-FedDyne,2020-FL-Scaffold}, we utilize the knowledge from clients' models to extract good knowledge for the aggregation model instead of using model parameters to reduce the physical distance between distributed models. Following that, our proposed framework can flexibly handle weight distance and probability distance in an efficient way, i.e., $\Vert p^k(y=c)-p(y=c) \Vert$ (please refer to Appendix~\ref{appendix:SGDissues}). 

\subsection{Multi-Teacher Knowledge Distillation}
\label{sec:multi-teacher-knowledge-distillation}
MTKD is an improved version of KD (which is presented in Appendix~\ref{appendix:knowledge-distillation}), in which multiple teachers work cooperatively to build a student model. As shown in \cite{2018-TL-KD-OnFly-NativeEnsemble}, every MTKD technique solves the following problem formulation: 
\begin{align}
    \textbf{P1}:\min \mathcal{L}_m^\textit{KL} 
    = \sum^R_{r=1} \sum^C_{l=1}\hat{p}(l|\boldsymbol{X},\boldsymbol{\omega}^r,T)\log{\frac{\hat{p}(l|\boldsymbol{X},\boldsymbol{\omega^r},T)}{\hat{p}(l|\boldsymbol{X},\boldsymbol{\omega}^g,T)}},
\label{eq:general-mtkd}
\end{align}
here, $r\in \{R\}$ are the teachers' indices. By minimizing \textbf{P1}, the student $\hat{p}^g$ can attain knowledge from all teachers. However, when using MTKD, there are some problems in extracting the knowledge distillation from multiple teachers. In particular, the process of distilling knowledge in MTKD is typically empirical without understanding the teacher's knowledge (i.e., aggregating all KL divergences between each teacher and the student). Therefore, MTKD is unable to exploit teachers' detailed predictions for the KD (e.g., \cite{2021-TL-MultiTeacher-MultiLevel}, \cite{2019-DL-EnsembleKD}, \cite{2018-TL-KD-OnFly-NativeEnsemble}, \cite{2017-DL-EnsembleTeachers}, \cite{2020-TL-HydraKD}). Another version of MTKD, KTMDs can only apply for a better teachers to distill knowledge (e.g., \cite{2019-DL-Customize-Student-HeterogenousTeachers}, \cite{2021-DL-Student-Customized-KD}, \cite{2022-DL-ConfidenceAware-KD}, \cite{2021-DL-DenselyGuidedKD}). For example, as provided in \cite[eq.~6]{2019-DL-Customize-Student-HeterogenousTeachers}, the student only selects the best teacher to operate the knowledge distillation. Visually, this technique is the same as the way of selecting a teacher among a set of teachers to carry out a single teacher distillation. Therefore, the student's performance is always bounded by the best teacher's performance. Another popular research direction in MTKD is to leverage the advantage of the gap between teachers' hidden class features. However, owing to the lack of explanatory knowledge in teachers' hidden layers, the method in \cite{2021-DL-Student-Customized-KD} cannot obtain better student performance when compared to their teachers. Generally, current MTKD techniques cannot extract good knowledge from different customer models, leading to weight divergence in FL. 

\section{Full-stack Federated Learning}
\label{sec:IV-ProposedAlgorithm}
\subsection{The F2L framework}
\label{sec:IV-C-MultiteacherKD}
The main objective of our work is to design a hierarchical FL framework, in which a global server manages a set of distinct regional servers. Utilizing hierarchical FL, our proposed algorithm can achieve computation and computation efficiency. The reason is that Hierarchical FL makes the clients to train sectionally before making the global aggregation \cite{2020-FL-Hierarchical, 2020-FL-HierarchicalClustering}. Consequently, FL inherits every advantage from mobile edge intelligence concept over traditional non-hierarchical networks (e.g., communication efficiency, scalability, controlability) \cite{2020-5G-Survey,2016-IoT-Survey,2020-FLMEC-Survey}. At the end of each knowledge-sharing episode, the regions (which are supervised by regional servers) cooperate and share their knowledge (each region functions as a distinguished FL system, with a set amount of communication rounds per episode).  

In each episode, each region randomly selects a group of clients from the network to carry out the aggregation process (e.g., FedAvg, FedProx); therefore, each region functions as a small-scale FL network. As a result, there are always biases in label-driven performance by applying random sampling on users per episode (see Appendix~\ref{appendix:proof-on-sampling-and-data-distribution}).
Given the random sampling technique applied to the regional client set, the regions always have different regional data distributions. 
Consequently, various label-driven performances of different teachers might be achieved.

At the global server, our goal is to extract good performance from regional teachers while maintaining the salient features (e.g., understanding of the regional data distributions) of all regions.
As a result, we can capture useful knowledge from diverse regions in each episode using our proposed innovative knowledge distillation technique (which is briefly demonstrated in Section~\ref{sec:IV-D-LabelDrivenKD}). We train the model on the standard dataset on the central server to extract knowledge from multiple teachers into the global student model. The preset data pool on the server $\mathcal{S}$ is used to verify the model-class reliability and generate pseudo labels. 

The system model is illustrated in Fig.~\ref{fig:hierarchicalFL}, and thoroughly described in Appendix~\ref{sec:system-model}.  
\begin{figure*}[t]
\centering
\includegraphics[width = 0.8\linewidth]{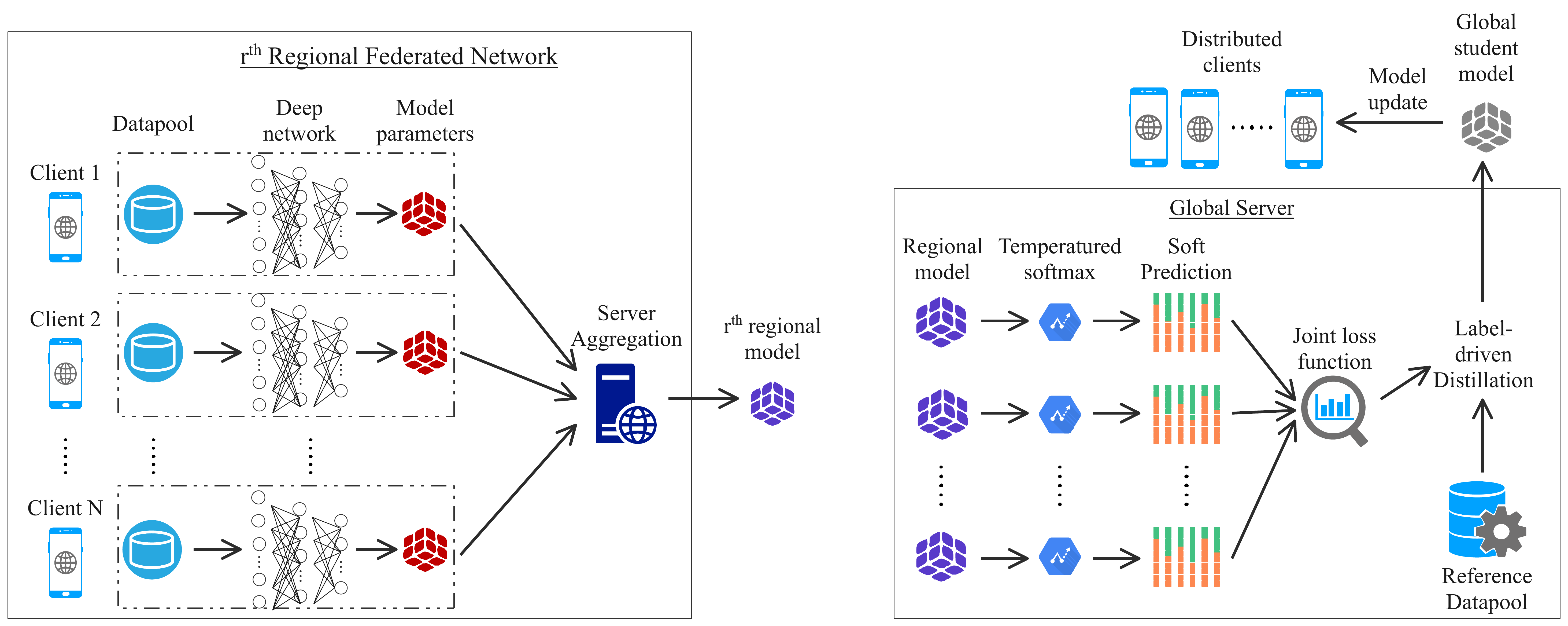}
\caption{The architecture of our F2L framework.}
\label{fig:hierarchicalFL}
\end{figure*}
The pseudo algorithm for F2L is demonstrated in Algorithm~\ref{alg:F2L}. When the FL process suffers from client-drift \cite{2020-FL-Scaffold} (i.e., the distribution of label-driven accuracies of different regions have large variance), the F2L framework applies LKD to reduce the class-wise performance gaps between regions (i.e., the regions with better performance on a particular class share their knowledge to regions with low performance). As a result, the FL network achieves a significantly faster convergence when utilizing LKD (which is intensively explained in Section~\ref{sec:IV-D-LabelDrivenKD}.) for the global aggregation process. When the generalization gap between regions is considerably reduced (i.e., $\Vert\max_r{\beta^c_r} - \min_r{\beta^c_r}\Vert \leq \epsilon$), our F2L network becomes vanilla FL to reduce computation and communication costs. To this end, our method can achieve computation efficiency while showing robustness to the non-IID data in the network. Additionally, whenever a new set of clients are added into the network and makes positive contributions to the FL system (e.g., $\Vert\max_r{\beta^c_r} - \min_r{\beta^c_r}\Vert \geq \epsilon$ where $\Vert\max_r{\beta^c_r}\Vert$ a corresponding to the new region's performance) the LKD aggregator can be switched back to improve the FL system's performance over again.


\subsection{Label-driven Knowledge Distillation}
\label{sec:IV-D-LabelDrivenKD}
To extract knowledge from multiple teachers to the global student model, we train the model on the standard dataset on the central server. The preset data pool on the server $\mathcal{S}$ is used to verify the model-class reliability and generate pseudo labels. In our work, the MTKD process executes two main tasks: (1) extracting the teachers' knowledge and (2) maintaining the students' previous performance.

To comprehend the LKD method, we first revisit the conventional MTKD, where the probabilistic output is calculated by model $\boldsymbol{\omega}$ on $x_i$, the prediction label $c$ is $\hat{p}(l|x_i,\boldsymbol{\omega},T,c)$ and its relation is: 
\begin{align}\label{LKD-surrogateOutput}
\hat{p}(l|x_i,\boldsymbol{\omega},T,c) = 
\begin{cases}
    \hat{p}(l|x_i,\boldsymbol{\omega},T), & \text{if } \text{argmax}\left[ \hat{p}(l|x_i,\boldsymbol{\omega},T)\right] = c, \\
    0, & \text{otherwise}.
\end{cases}
\end{align}

On the one hand, we aim to transfer knowledge from different regional models to the global one. Inspired by \cite{2015-DL-KnowledgeDistillation}, we use the Kullback–Leibler (KL) divergence between each regional teacher and the global logits as a method to estimate the difference between two models' performance. The relationship is expressed as follows:
\begin{align}\label{LKD-TeaStu-Loss-Weighted}
\mathcal{L}_r^\textit{KL} = \sum^C_{c=1} &\beta^c_r  \sum^{S^r_c}_{i=1} \sum^C_{l=1} \hat{p}^r(l|x_i,\boldsymbol{\omega}^r,T,c) 
\times\log{\frac{\hat{p}^r(l|x_i,\boldsymbol{\omega}^r,T,c)}{\hat{p}^g(l|x_i,\boldsymbol{\omega}^g,T,c)}},
\end{align}
where $S$ is the number of samples of the fixed dataset $\mathcal{S}$ on the server. $(\boldsymbol{X}^r_\text{alg}, \boldsymbol{Y}^r_\text{alg})$ is the dataset which is pseudo labeled and aligned by regional model $r$ and $(\boldsymbol{X}^r_\text{alg}[c], \boldsymbol{Y}^r_\text{alg}[c])$ represents the set of data with size of $S_c^r$ labeled by the model $r$ as $c$. Although the same preset dataset is utilized on every teacher model, the different pseudo labeling judgments from different teachers lead to the different dataset tuples. The process of identifying $S_c^r$ is demonstrated in Algorithm~\ref{alg:L-SampleAlign}. Because the regional models label on the same dataset $S$, we have $\sum_{c=1}^C S^r_c = S$ for all regional models. $D_\textit{KL}^c(\hat{p}^r||\hat{p}^g)$ is the $c$ label-driven KL divergence between model $r$ and model $g$. 

On the other hand, we aim to guarantee that the updated global model does not forget the crucial characteristics of the old global model. Hence, to measure the divergence between the old and the updated model, we introduce the following equation: 
\begin{align}\label{LKD-OldNew-Loss-Weighted}
\mathcal{L}_{\boldsymbol{\omega}_\textit{upd}}^\textit{KL}
= \sum^C_{c=1} &\beta^c_{\boldsymbol{\omega}_\textit{old}}  \sum^{S^r_c}_{i=1} \sum^C_{l=1} \hat{p}^g(l|x_i,\boldsymbol{\omega}^g_\textit{old},T,c) 
\times\log{\frac{\hat{p}^g(l|x_i,\boldsymbol{\omega}^g_\textit{old},T,c)}{\hat{p}^g(l|x_i,\boldsymbol{\omega}^g_\textit{new},T,c)}},
\end{align}
where $\boldsymbol{\omega}_\textit{old}$ is the old parameters set of the global model which is distilled in the last episode of F2L. More details about the label-driven knowledge distillation are discussed in Appendix~\ref{appendix:labeldriven-knowledgedistillation}.

To compare the performance between LKD and MTKD, we consider the following assumption and lemmas: 
\begin{lemma}
Given $\tau^c_r$ is the $c$-label driven predicting accuracy on model $r$. Let $\sigma^2_{r,c}, \mu_{r,c}$ be the model's variance and mean, respectively. The optimal value of variance and mean on student model (i) $\sigma^{*2}_{LKD,g,c}, \mu^*_{LKD,g,c}$ yields $\sigma^{*2}_{\textrm{LKD},g,c} = \frac{1}{\sum_{r=1}^{R} e^{\tau^c_r}} \sum^R_{r=1} e^{\tau_r^c}\sigma^2_{r,c}$, and $\mu^{*}_{\textrm{LKD},g,c} = \frac{1}{\sum_{r=1}^{R} e^{\tau^c_r}} \sum^R_{r=1} e^{\tau_r^c}\mu_{r,c}.$. 
\label{lemma:optimal-LKD}
\end{lemma}
\textit{Proof:} The proof is provided in Appendix~\ref{appendix:lemma-1}.

\begin{assumption}
    Without loss of generality, we consider  $R$ distinct regional models whose accuracy satisfy the following prerequisites $\sigma^2_{1,c} \leq \sigma^2_{2,c} \leq \ldots \leq \sigma^2_{R,c}$, and $|\mu_{1,c} - \Bar{\mu}_c| \leq |\mu_{2,c} - \Bar{\mu}_c| \leq \ldots \leq |\mu_{R,c} - \Bar{\mu}_c|$ ($\Bar{\mu}_c$ is denoted as an empirical global mean of the dataset on class $c$).
\end{assumption}

\begin{lemma}
Given the set of models with variance satisfy $\sigma^2_{1,c} \leq \sigma^2_{2,c} \leq \ldots \leq \sigma^2_{R,c}$, the models' accuracy have the following relationship $\tau^c_1 \geq \tau^c_2 \geq \ldots \geq \tau^c_R$.
\label{lemma:relationship-accuracy-variance}
\end{lemma}
\textit{Proof.} The proof can be found in Appendix~\ref{appendix:relationship-accuracy-variance}.

\begin{theorem}
Let $\sigma^{*2}_{\textrm{LKD},g,c} $ be the class-wise variance of the student model, and $\sigma^{*2}_{\textrm{MTKD},g,c}$ be the class-wise variance of the model of teacher $r$, respectively. We always have the student's variance using LKD technique always lower than that using MTKD:
\begin{align}
    \sigma^{*2}_{\textrm{LKD},g,c} 
    \leq 
    \sigma^{*2}_{\textrm{MTKD},g,c}.
\label{eq:LKD-vs-MTKD}
\end{align}
\label{theorem:labeldriven-knowledgedistillation-analysis}
\end{theorem}
\textit{Proof}: For the complete proof see Appendix \ref{appendix:labeldriven-knowledgedistillation-analysis}.
\begin{theorem}
Let $\mu^{*}_{\textrm{LKD},g,c} $ be the empirical $c$-class-wise mean of the student model, and $\mu^{*}_{\textrm{MTKD},g,c}$ be the empirical $c$-class-wise mean of the model of teacher $r$, respectively. We always have the student's empirical mean using LKD technique always closer to the empirical global dataset's class-wise mean ($\Bar{\mu}_c$) than that using MTKD:
\begin{align}
    |\mu^{*}_{\textrm{LKD},g,c} - \Bar{\mu}_c|
    \leq 
    |\mu^{*}_{\textrm{MTKD},g,c} - \Bar{\mu}_c|.
\label{eq:LKD-vs-MTKD-mean}
\end{align}
\label{theorem:labeldriven-knowledgedistillation-analysis-mean}
\end{theorem}
Given Theorems~\ref{theorem:labeldriven-knowledgedistillation-analysis} and \ref{theorem:labeldriven-knowledgedistillation-analysis-mean}, we can prove that our proposed LKD technique can consistently achieve better performance than that of the conventional MTKD technique. Moreover, by choosing the appropriate LKD allocation weights, we can further improve the LKD performance over MTKD. 
Due to space limitation, we defer the proof to Appendix \ref{appendix:labeldriven-knowledgedistillation-analysis-mean}.

\subsection{Class Reliability Scoring}
The main idea of class reliability variables $\beta^c_r$, $\beta^c_{\boldsymbol{\omega}_\textit{old}}$ in LKD is to weigh the critical intensity of the specific model. Therefore, we leverage the attention design from \cite{2017-DL-Attention} to improve the performance analysis of teachers' label-driven.

For regional models with disequilibrium or non-IID data, the teachers only teach the predictions relying upon their specialization. The prediction's reliability can be estimated by leveraging the validation dataset on the server and using the function under the curve (AUC) as follows: 
\begin{equation}\label{LKD-teacher_weight}
\begin{split} 
\beta^c_r = \frac{\text{exp}(f_\textit{AUC}^{c,r} T_{\boldsymbol{\omega}})}{\sum^R_{r=1}\text{exp}(f_\textit{AUC}^{c,r} T_{\boldsymbol{\omega}} )},
\end{split}
\end{equation}
where $f^{c,r}_\text{AUC}$ denotes the AUC function on classifier $c$ of the regional model $r$. Since AUC provides the certainty that a specific classifier can work on a label over the rest, we use the surrogate softmax function to weigh the co-reliability among the same labeling classifiers on different teacher models. For simplicity, we denote $\beta^c_{\boldsymbol{\omega}_\textit{old}}$ as the AUC on each labeling classifier: 
\begin{equation}\label{LKD-oldModel_weight}
\begin{split} 
\beta^c_{\boldsymbol{\omega}_\textit{old}} = \frac{\text{exp}(f_\textit{AUC}^{c,\boldsymbol{\omega}_\text{old}} T_{\boldsymbol{\omega}})}{\text{exp}(f_\textit{AUC}^{c,\boldsymbol{\omega}_\text{new}} T_{\boldsymbol{\omega}})+\text{exp}(f_\textit{AUC}^{c,\boldsymbol{\omega}_\text{old}} T_{\boldsymbol{\omega}} )}.
\end{split}
\end{equation}

In the model update class reliability, instead of calculating the co-reliability between teachers, \eqref{LKD-oldModel_weight} compares the performance of the previous and current global models. Moreover, we introduce a temperatured value for the class reliability scoring function, denoted as $T_{\boldsymbol{\omega}}$. By applying a large temperatured value, the class reliability variable sets $\beta^c_r$, and $\beta^c_{\boldsymbol{\omega}_\textit{old}}$ make a higher priority on the better performance (i.e., the label-driven performance on class $c$ from teacher $r$, e.g., $f_\textit{AUC}^{c,r}$ in equation~\eqref{LKD-teacher_weight} or class $c$ from old model $\boldsymbol{\omega}_\text{old}$ in equation~\eqref{LKD-oldModel_weight}). By this way, we can preserve the useful knowledge which is likely ignored in the new distillation episode. The more detailed descriptions of class reliability scoring are demonstrated in Algorithm~\ref{alg:C-Reliability}.
\subsection{Joint Multi-teacher Distillation for F2L}
We obtain the overall loss function for online distillation training by the proposed F2L:
\begin{align}
\mathcal{L}_\text{F2L} = \lambda_1 \sum^R_{r=1}\mathcal{L}^\textit{KL}_r + \lambda_2\mathcal{L}^{KL}_{\boldsymbol{\omega}_\textit{upd}} + \lambda_3\mathcal{L}^g_\textit{CE},
\end{align}
where $\lambda_1, \lambda_2, \lambda_3$ are the scaling coefficients of the three terms in the joint loss function. The first and second terms imply the joint LKD from the regional teacher models and the updating correction step, respectively. Moreover, to ensure that knowledge the student receives from teachers is accurate and can be predicted accurately in practice, we need to validate the quality of the student model on the real data. Thus, we also compute the “standard” loss between the student and the ground-truth labels of the train dataset. This part is also known as the hard estimator, which is different from the aforementioned soft-distillator. The hard loss equation is as follows:
\begin{align}
\mathcal{L}^g_\textit{CE} &= H(y,\hat{p}(l|\boldsymbol{X},\boldsymbol{\omega}^g,T))
= \sum^C_{l=1} y_l \log{\hat{p}(l|\boldsymbol{X},\boldsymbol{\omega}^g,T)}.
\end{align}
\begin{algorithm}[h]
    \caption{F2L framework}
    \label{alg:F2L}
\begin{algorithmic}
    \State {\bfseries Require:} Initialize clients' weights, global aggregation round, number of regions $R$, arbitrary $\epsilon$.
    \While{not converge}
        \For{all regions $r\in \{1,2,\dots,R\}$}
        \For{all user in regions}
        \State Apply FedAvg on regions $r$.
        \EndFor
        \State Send regional model $\boldsymbol{\omega}^r$ to the global server.
        \EndFor
        \If{reach global aggregation round}
            \If{$\Vert\max_r{\beta^c_r} - \min_r{\beta^c_r}\Vert \geq \epsilon$ where $\beta = \{\beta^1_r,\dots,\beta^C_r\}\vert^R_{r=1}$ from Algorithm~\ref{alg:C-Reliability}}
                \State Apply LKD as described in Algorithm~\ref{alg:LD-KD}
            \Else 
                \State $\boldsymbol{\omega}^g = 1/R \sum^R_{r=1} \boldsymbol{\omega}^r$. 
            \EndIf 
        \EndIf
    \EndWhile
\end{algorithmic}
\end{algorithm}

We use the temperature coefficient $T = 1$ to calculate the class probability for this hard loss. The overall training algorithm for LKD is illustrated in Algorithm~\ref{alg:LD-KD}. In terms of value selection for scaling coefficients, the old global model can be considered as an additional regional teacher's model in the same manner, in theory. Therefore, $\lambda_2$ should be chosen as:
\begin{align}\label{LKD-lambda2-choice}
\lambda_2 = 
\begin{cases}
    \frac{1}{R}\lambda_1, & \text{if update distillation in \eqref{LKD-OldNew-Loss-Weighted} is considered,}  \\
    0, & \text{otherwise},
\end{cases}
\end{align}
where $R$ is the number of regions decided by our hierarchical FL settings. With respect to $\lambda_3$, the value is always set as: 
\begin{align}\label{LKD-lambda1-choice}
\lambda_3 = 
\begin{cases}
    1 - \frac{R+1}{R}\lambda_1, & \text{if update distillation in \eqref{LKD-OldNew-Loss-Weighted} is considered,}  \\
    1 - \lambda_1, & \text{otherwise}.
\end{cases}
\end{align}

\subsection{Discussions on the extent of protecting privacy} 
In its simplest version, our proposed F2L framework, like the majority of existing FL approaches, necessitates the exchange of models between the server and each client, which may result in privacy leakage due to, for example, memorization present in the models.
Several existing protection methods can be added to our system in order to safeguard clients against enemies. These include adding differential privacy \cite{2017-FL-DifferentiallyPrivate} to client models and executing hierarchical and decentralized model fusion by synchronizing locally inferred logits, for example on random public data, as in work \cite{2019-FL-Cronus}. We reserve further research on this topic for the future.

\section{Experimental Evaluation}
\label{sec:V-ExperimentalEvaluation}
\subsection{Comparison with FL methods}
\label{sec:V-B-Comparison}
\begin{table*}[t]
\addtolength{\tabcolsep}{-3pt}
\centering
\caption{The top-1 test accuracy of different baselines on different data settings. The $\alpha$ indicates the non-IID degree of the dataset (the lower value of $\alpha$ means that the data is more heterogeneous).\\}
\begin{tabular}{|c|c|c|c|c|c|c|c|c|c|c|}
\hline
Dataset   & FedAvg & FedGen & FedProx & Fed-    & F2L    & FedAvg & FedGen & FedProx & Fed-    & F2L    \\
          &        &        &         & Distill & (Ours) &        &        &         & Distill & (Ours) \\ 
\hline \hline
\cline{1-11}                      
                       & \multicolumn{5}{|c|}{Dirichlet ($\alpha=1$)} & \multicolumn{5}{|c|}{Dirichlet ($\alpha=0.1$)}                                                \\
\cline{1-11}                      
                       EMNIST    & $71.66$& $78.70$& $70.77$ & $75.56$& $\textbf{81.14}$ & $59.10$    & $68.24$& $58.88$ & $46.03$  & $\textbf{68.31}$ \\
                       CIFAR-10  & $60.48$& $59.21$& $63.72$ & $62.36$& $\mathbf{71.22}$ & $47.07$& $47.08$& $47.05$ & $45.67$  & $\mathbf{55.22}$ \\
                       CIFAR-100 & $36.17$& $40.26$ & $36.3$   & $34.88$   & $\mathbf{50.33}$ & $21.31$& $28.96$& $20.43$ & $16.15$ & $\mathbf{31.07}$ \\
                       CINIC-10 & $65.23$& $71.61$& $65.15$ & $67.77$ & $\mathbf{74.85}$ & $47.55$& $52.35$& $48.2$ & $47.1$ & $\mathbf{57.12}$\\
                       CelebA & $70.82$& $75.43$& $71.07$ & $68.59$ & $\mathbf{81.65}$   & $63.58$& $70.14$& $66.33$ & $62.91$ & $\mathbf{74.14}$\\                
\hline
\end{tabular}
\label{tab:F2L-Comparisons} 
\end{table*}
We run the baselines (see Section~\ref{sec:V-A-ExperimentSetup}) and compare with our F2L. Then, we evaluate the comparisons under different non-IID ratio. More precisely, we generate the IID data and non-IID data with two different Dirichlet balance ratio: $\alpha = \{1, 10\}$. The comparison results are presented in Table~\ref{tab:F2L-Comparisons}. As shown in Table~\ref{tab:F2L-Comparisons}, the F2L can outperform the four baselines with a significant increase in accuracy. The reason for this phenomenon is that the LKD technique selectively draws the good features from regional models to build a global model. Hence, the global model predicts the better result on each different class and the entire accuracy of the global model then increases tremendously. The significant impact when applying LKD to distill different teachers to one student is shown in Table~\ref{tab:LKD-Distillation-Efficiency}. 
\subsection{Computation Efficiency of F2L}
\label{sec:V-C-F2L-CommCompEfficiency}
\begin{figure}[!h]
	\centering
	\subfigure[\label{fig:F2L-performance}]{\includegraphics[width=0.3\linewidth]{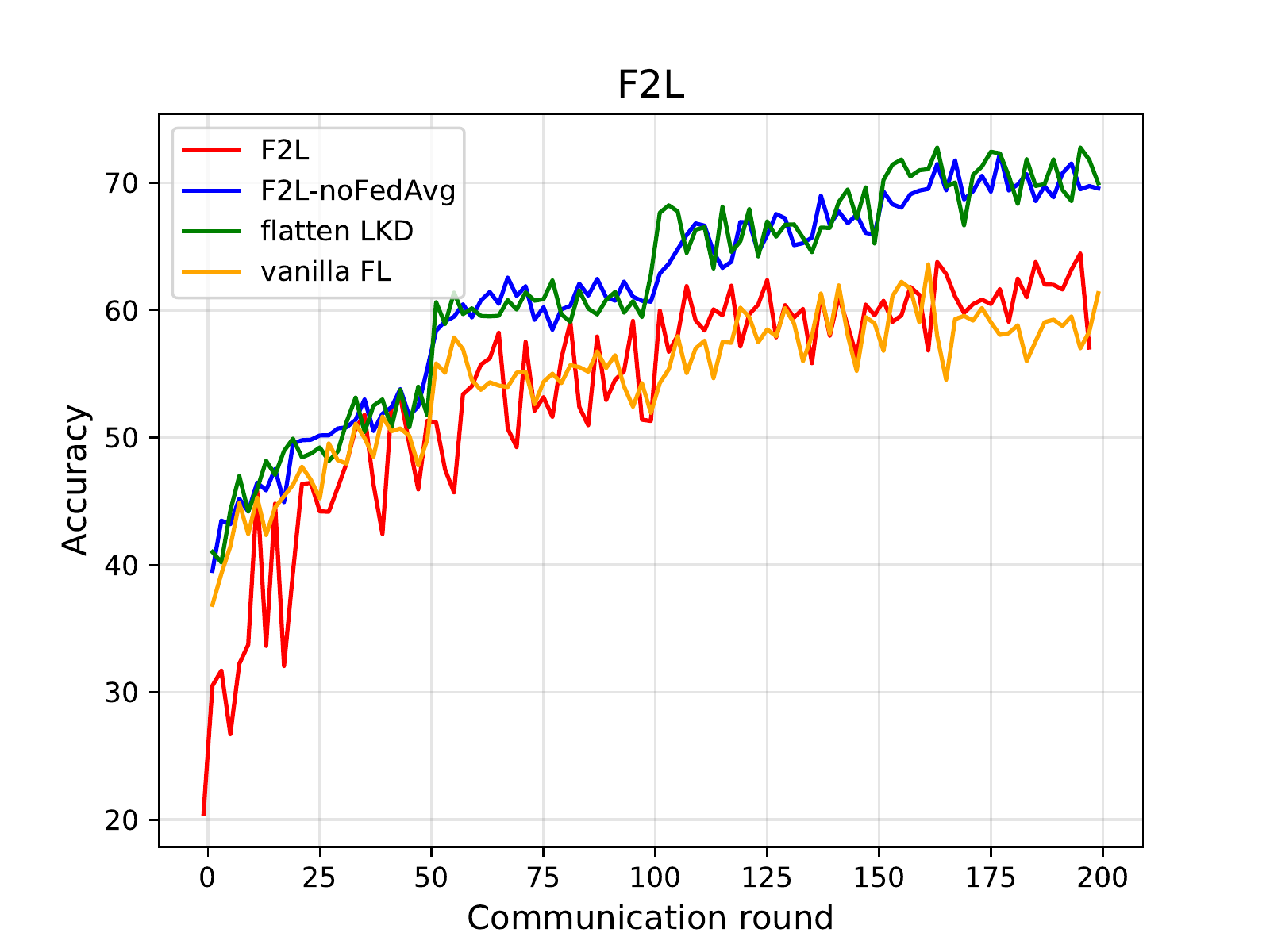}}
	\subfigure[\label{fig:F2L-compcost}]{\includegraphics[width=0.25\linewidth]{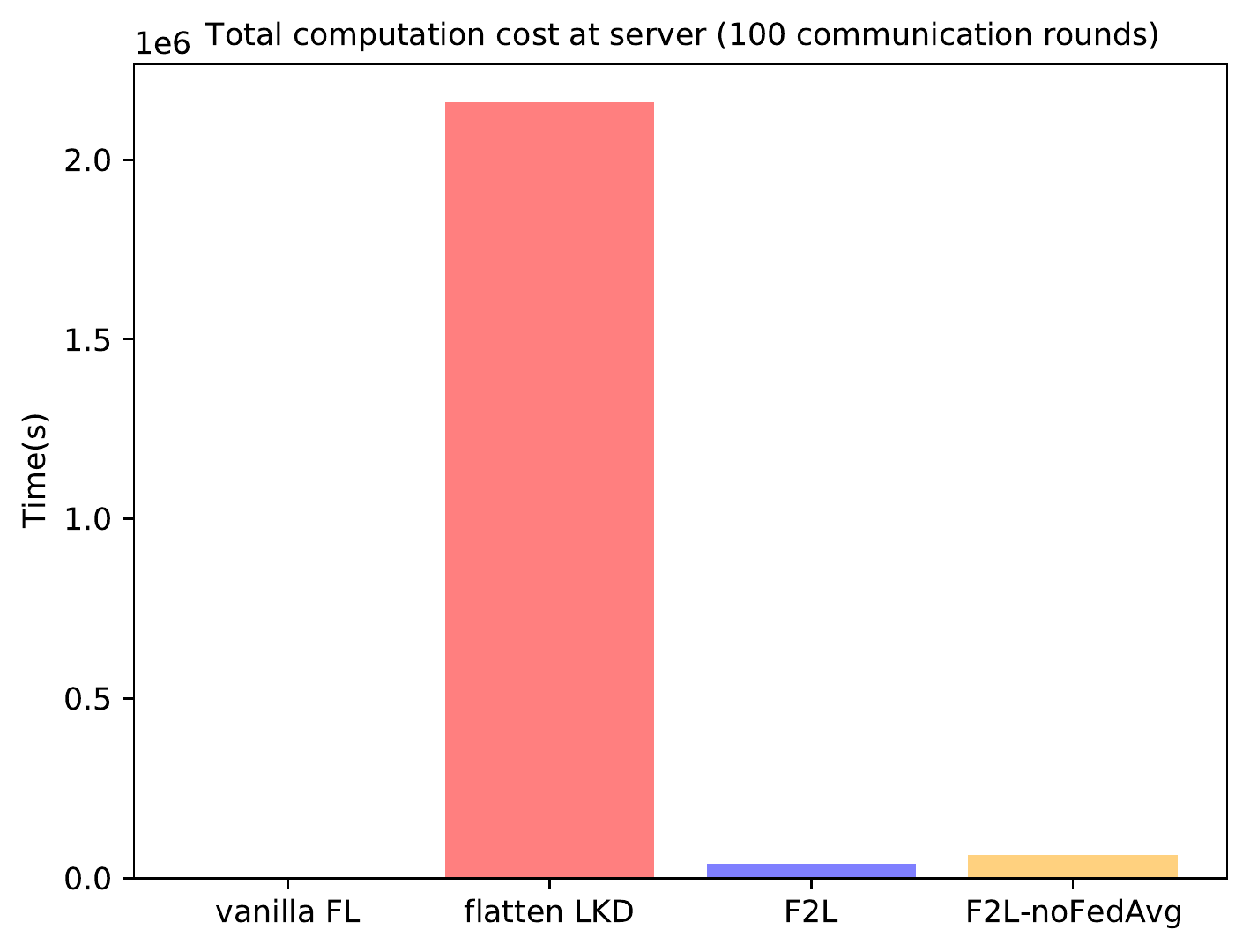}}
	\subfigure[\label{fig:F2L-scalability}]{\includegraphics[width=0.3\linewidth]{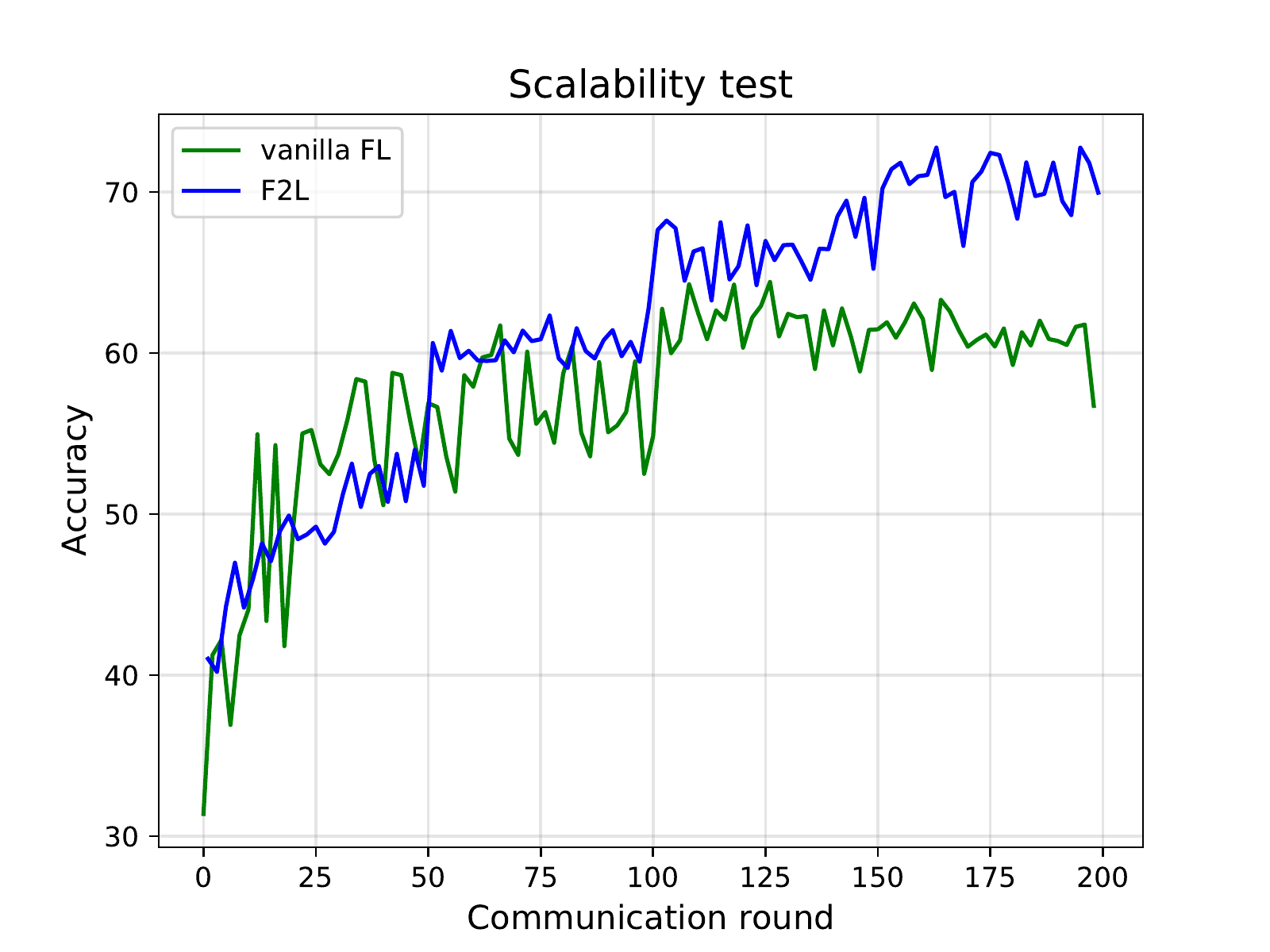}}
	\caption{Performance benchmarks of F2L under different settings. Fig.~\ref{fig:F2L-performance} reveals the convergence. Fig.~\ref{fig:F2L-compcost} shows the computational cost, and Fig.~\ref{fig:F2L-scalability} demonstrates the F2L convergence when a new set of clients are added into the FL system (i.e., at communication round $100$).}
	\label{fig:F2Ltest}
\end{figure}

To evaluate the computation efficiency of our proposed F2L process, we compare our F2L process with 3 benchmarks: (i) F2L-noFedAvg (aggregator only consists of LKD), (ii) vanilla FL (FL with flatten architecture and FedAvg as an aggregator), and (iii) flatten LKD (FL with flatten architecture based with LKD as an alternate aggregator). Fig.~\ref{fig:F2L-performance} shows that the F2L system can achieve convergence as good as the F2L-noFedAvg. The reason is that: after several communication rounds, the distributional distance between regions is reduced thanks to the LKD technique. Hence, the efficiency of the LKD technique on the data is decreased. As a consequence, the LKD technique shows no significant robustness over FedAvg aggregator. In the non-hierarchical settings, the flatten LKD and FedAvg reveal under-performed compared to the proposed hierarchical settings. We assume that the underperformance above comes from the data shortage of clients' training models. To be more detailed, the clients' dataset are considerably smaller than that of the \say{regional dataset}. Thus, the regional models contain more information than the clients' models. We believe that: in the LKD technique, teachers' models require a certain amount of knowledge to properly train a good student (i.e., the global model).
Given the convergence rate from Fig.~\ref{fig:F2L-performance} and the computation cost at the server on Fig.~\ref{fig:F2L-compcost}, we can see that, by using the adaptive switch between LKD and FedAvg in F2L, we can achieve significant computational efficiency at the aggregation server. Note that F2L can substantially increase performance and computation efficiency compared with non-hierarchical architecture.

\subsection{Scalability}
\label{sec:V-C-F2L-Scalability}
This section evaluates the F2L scalability. To do so, we inject a group of clients with  non-IID data into our FL system after $100$ rounds (when the convergence becomes stable). Note that the FL system has never trained these data. The detailed configurations of our experimental assessments can be found in Appendix~\ref{sec:V-A-ExperimentSetup}. As it can be seen from Fig.~\ref{fig:F2L-scalability}, when a new group of clients are added to the FL system, the vanilla FL shows a significant drop in terms of convergence. The reason is because of the distribution gap between the global model's knowledge and knowledge of the clients' data. Whenever new data with unlearned distribution is added to a stable model, the model will make considerable gradient steps towards the new data distribution. Thus, the FedAvg takes considerable learning steps to become stable again. In contrast, in F2L system, the learning from newly injected regions does not directly affect the learning of the whole FL system. Instead, the knowledge from the new domains is selectively chosen via the LKD approach. Thus, the LKD process does not suffer from information loss when new clients with non-IID data are added to the FL system.

\subsection{LKD Analysis}
\label{sec:V-C-LKD-Analysis}
In this section, we evaluate the LKD under various settings to justify the capability of LKD to educate the good student from the normal teachers. Our evaluations are as follows.

\begin{figure}[!h]
	\centering
	\subfigure[\label{fig:teacher1-prior-distill-evaluation}]{\includegraphics[width=0.235\linewidth]{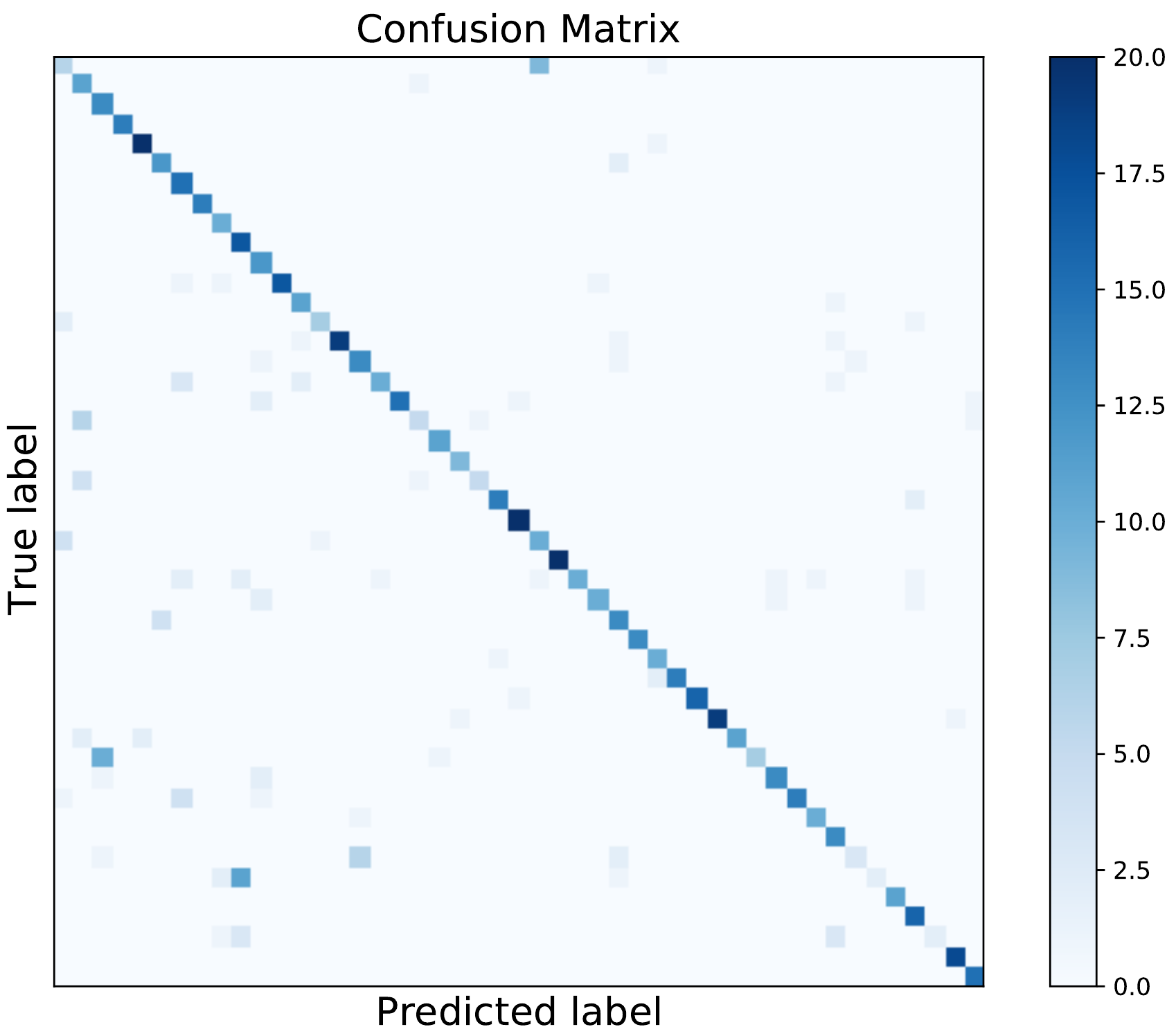}}
	\subfigure[\label{fig:teacher2-prior-distill-evaluation}]{\includegraphics[width=0.235\linewidth]{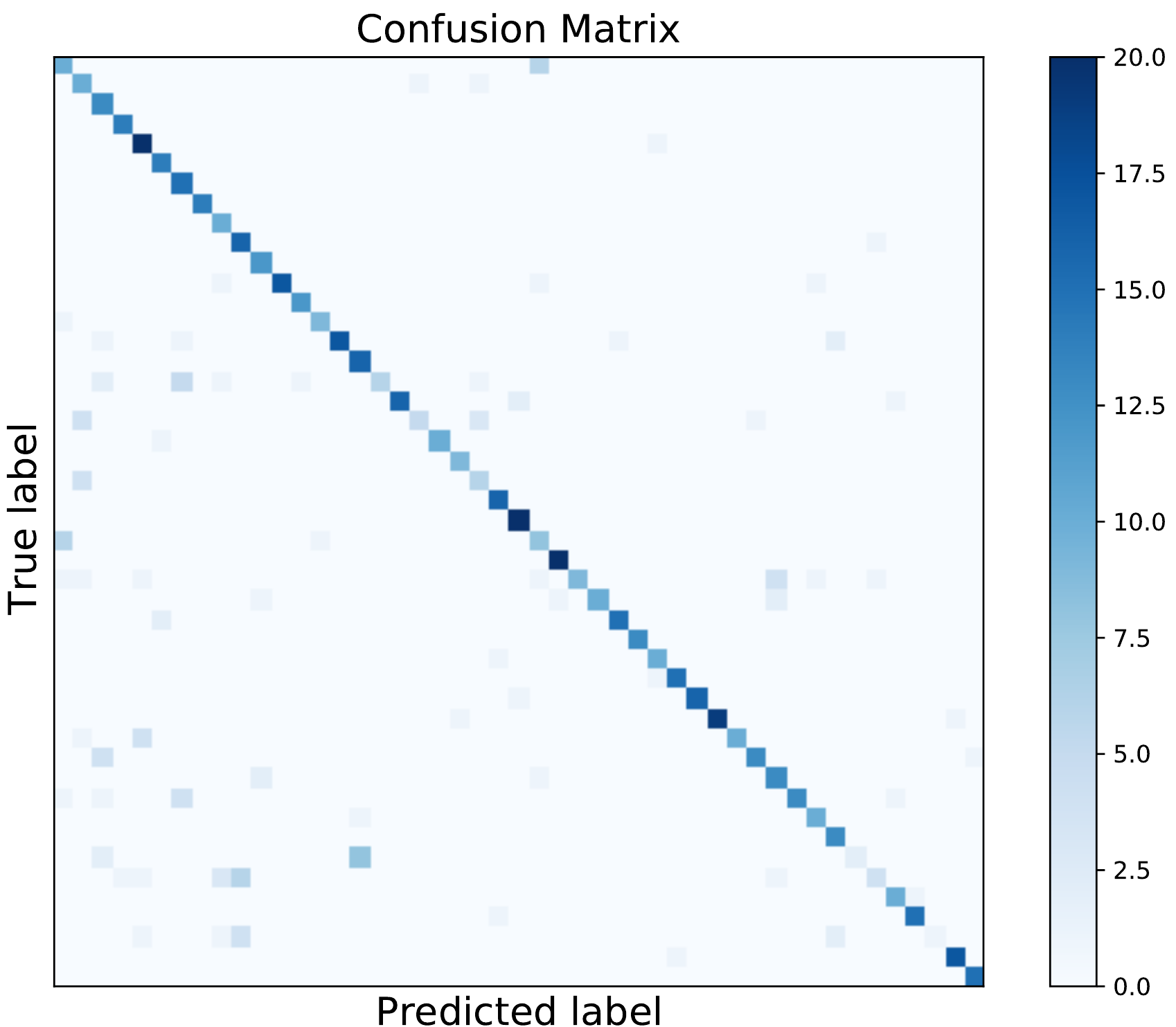}}
	\subfigure[\label{fig:teacher3-prior-distill-evaluation}]{\includegraphics[width=0.235\linewidth]{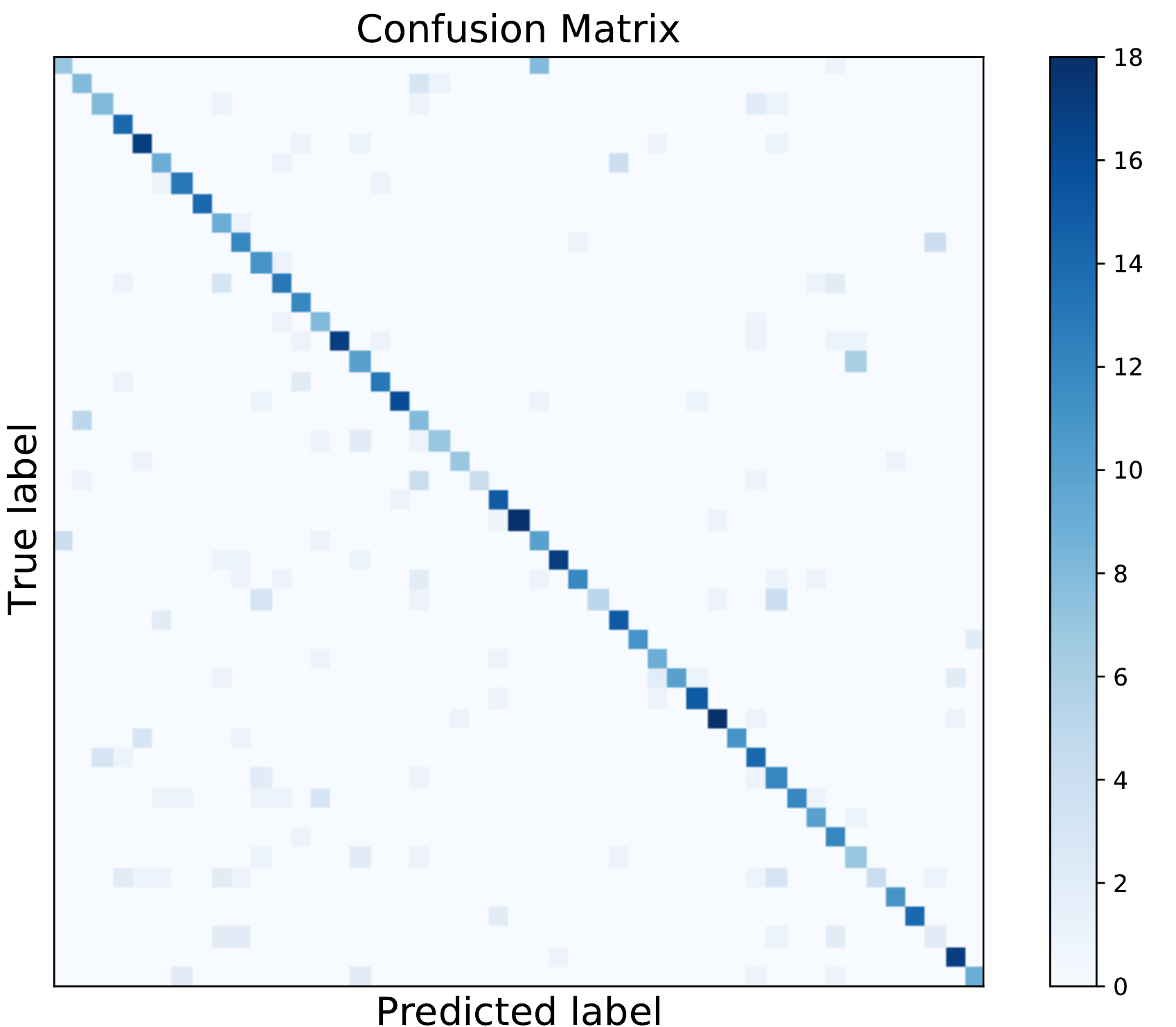}}
	\subfigure[\label{fig:student-later-distill-evaluation}]{\includegraphics[width=0.235\linewidth]{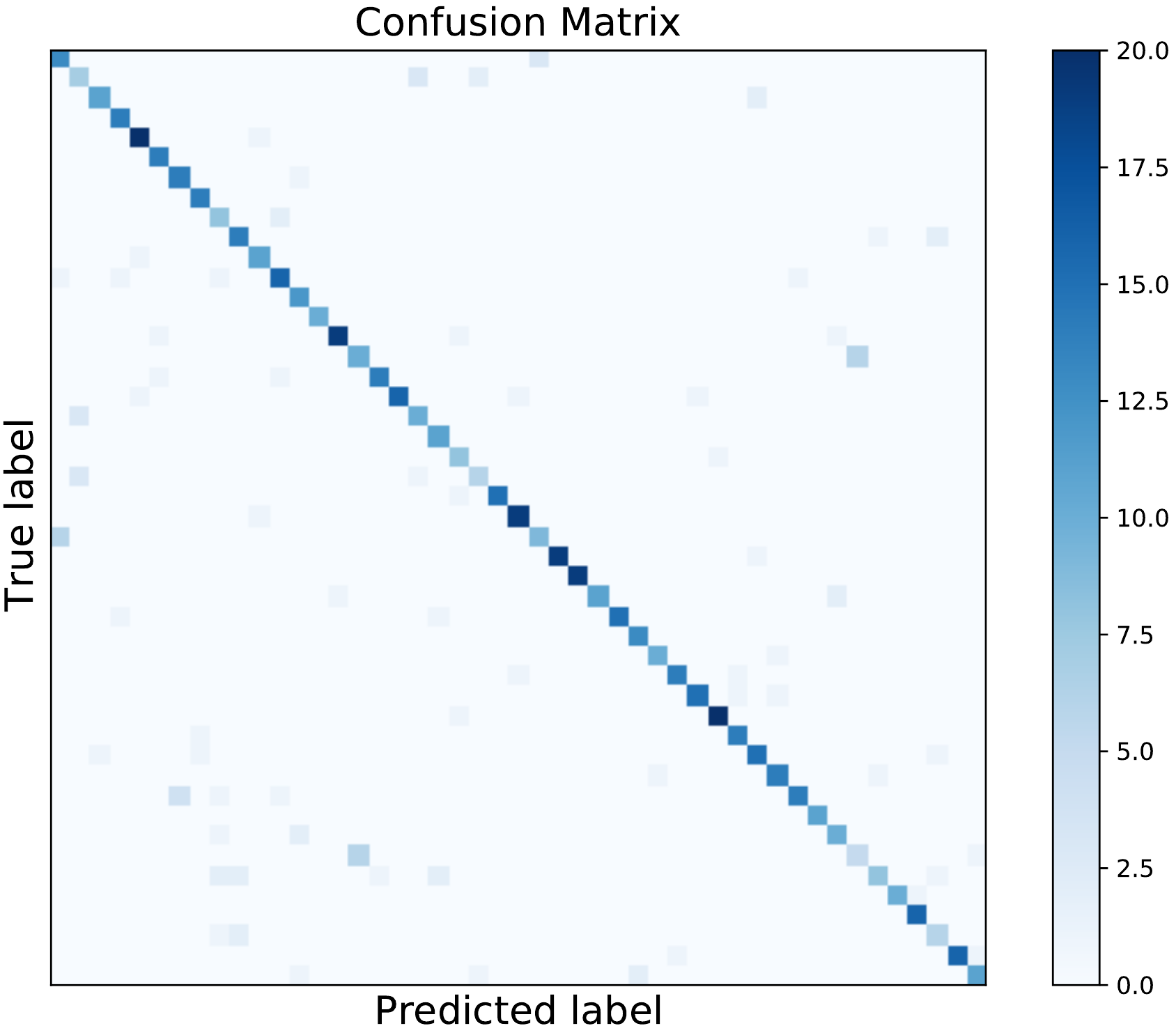}}
	\caption{The illustrative results of LKD on EMNIST dataset. Confusion matrices show the effectiveness of joint distillation on regional models. Figures (a), (b), and (c) are the confusion matrix before distillation of teacher's predictions in region $1$, $2$, and $3$, respectively (see Appendix~\ref{sec:V-A-ExperimentSetup}). Fig.~(d) is the confusion matrix of predictions after distillation of student. The matrix diagonal demonstrates the true-predicted label of the model.}
	\label{fig:ConfusionMatrix}
\end{figure}

\textbf{Can student outperform teachers?}
To verify the efficiency of LKD with respect to enhancing student performance, we first evaluate F2L on MNIST, EMNIST, CIFAR-$100$, CINIC-$10$, CelebA dataset. The regions are randomly sampled from the massive FL network. In this way, we only evaluate the performance of LKD on random teachers. Table~\ref{tab:LKD-Distillation-Efficiency} shows top-1 accuracy on the regional teacher and student models. The results reveal that LKD can significantly increase the global model performance compared with that of the regional models. Moreover, the newly distilled model can work well under each regional non-IID data after applying the model update. 
\begin{table}[t]
\renewcommand{\arraystretch}{1.25}
\caption{Top-1 accuracy of F2L on $5$ datasets MNIST, EMNIST, CIFAR-100, , CINIC-$10$ and CelebA. The data's heterogeneity is set at $\alpha = 0.1$ on CIFAR-100, MNIST, CINIC-$10$ and CelebA. We use EMNIST ``unbalanced'' to evaluate in this test. The ``before update'' and ``after update'' denote the teacher models' accuracies before and after the global distillation, respectively.} 
\centering 
\small\addtolength{\tabcolsep}{-3pt}
\begin{tabular}{|l|c|c|c|c|c|c|c|c|c|c|}
\hline
\multirow{3}{*}{\begin{tabular}[c]{@{}c@{}} \\ \end{tabular}} &
\multicolumn{2}{c|}{MNIST} 
                  & \multicolumn{2}{c|}{EMNIST} 
                  & \multicolumn{2}{c|}{CIFAR-100} 
                  & \multicolumn{2}{c|}{CINIC-10} 
                  & \multicolumn{2}{c|}{Celeb-A} \\ \cline{2-11} 
               &  before  & after  & before & after & before & after & before & after & before & after  \\ \hline
               &  update  & update  & update & update & update & update & update & update & update & update    \\ \hline
Teacher 1      & $61.02$  & $\mathbf{95.19}$ & $73.27$ & $\mathbf{84.09}$ & $20.11$ & $\mathbf{35.41}$ & $43.8$ & $\mathbf{46.59}$ & $62.37$ & $\mathbf{67.98}$ \\  \hline
Teacher 2      & $92.49$  & $\mathbf{98.22}$ & $78.80$ & $\mathbf{83.62}$ & $18.82$ & $\mathbf{31.2}$ & $42.15$ & $\mathbf{46.01}$ & $63.79$ & $\mathbf{72.33}$  \\  \hline
Teacher 3      & $81.60$  & $\mathbf{97.63}$ & $80.5$  & $\mathbf{84.10}$ & $22.40$ & $\mathbf{34.93}$ & $40.02$ & $\mathbf{42.15}$ & $64.05$ & $\mathbf{69.44}$ \\  \hline
G-student      & $\mathbf{98.71}$  &         & $\mathbf{84.11}$ &         & $\mathbf{37.68}$ &    & $\mathbf{47.65}$ &  & $\mathbf{70.12}$ &  \\ \hline
\end{tabular}
\label{tab:LKD-Distillation-Efficiency} 
\end{table}
\newline
To make a better visualization for the LKD's performance, we reveal the result of LKD on EMNIST dataset in terms of confusion matrix as in Fig.~\ref{fig:ConfusionMatrix}. As it can be seen from the figure, the true predictions is represented by a diagonals of the matrices. A LKD performance is assumed to be well predicted when the value on diagonals is high (i.e., the diagonals' colors is darker), and the off-diagonals is low (i.e., the off-diagonals' colors are lighter). As we can see from the four figures, there are a significant reduce in the off-diagonals' darkness in the student performance (i.e., Fig.~\ref{fig:student-later-distill-evaluation}) comparing to the results in other teachers (i.e., Figures~\ref{fig:teacher1-prior-distill-evaluation}, \ref{fig:teacher2-prior-distill-evaluation}, and \ref{fig:teacher3-prior-distill-evaluation}). Therefore, we can conclude that our proposed MTKD techniques can surpass the teachers' performance as we mentioned in Section~\ref{sec:related-works}.

\textbf{Teachers can really educate student?}
We evaluate LKD under different soft-loss coefficients $\lambda_1$ while the hard-loss factor is set at $\lambda_3 = 1-\lambda_1$ (the scaling value $\lambda_2$ is set to $0$). Thus, we can justify whether the robust performance of LKD comes from the joint distillation from teachers or just the exploitation of data-on-server training. We evaluate LKD on six scaling values $\lambda_1 = \{0, 0.001, 0.01, 0.1, 0.5, 1\}$. We evaluate on three dataset, including EMNIST, CIFAR-10, and CIFAR-100, and summarize the results in Tables~\ref{tab:LKD-DistillationTest-EMNIST}, \ref{tab:LKD-DistillationTest-CIFAR10} and \ref{tab:LKD-DistillationTest-CIFAR100} in \textbf{Appendices}. We can see from the three tables that the LKD cap off with $\lambda_3=0.01$. Especially, when $\lambda_3=1$ (which means the LKD acts as a vanilla cross-entropy optimizer), the model accuracy reduces notably. This means that the LKD only uses hard-loss as a backing force to aid the distillation. Thus, our LKD is appropriate and technically implemented.

\textbf{Required training sample size for joint distillation.} To justify the ability of LKD under a shortage of training data, we evaluate LKD with six different data-on-server settings: $\sigma = \{1, 1/2, 1/4, 1/6, 1/8, 1/10\}$, where $\sigma$ is the sample ratio when compared with the original data-on-server as demonstrated in Table~\ref{tab:data-configuration}. As we can see from the implementation results in three Tables~\ref{tab:LKD-SampleTrainTest-EMNIST}, \ref{tab:LKD-SampleTrainTest-CIFAR10}, and \ref{tab:LKD-SampleTrainTest-CIFAR100} in \textbf{Appendices}, the F2L is demonstrated to perform well under a relatively small data-on-server. To be more specific, we only need the data-on-server to be $4$ times lower than the average data-on-client to achieve a robust performance compared with the vanilla FedAvg. However, we suggest using the data-on-server to be larger than the data from distributed clients to earn the highest performance for LKD. Moreover, due to the ability to work under unlabeled data, the data-on-server does not need to be all labeled. We only need a small amount of labeled data to aid the hard-loss optimizer. Thus, the distillation data on the server can be updated from distributed clients gradually.
\section*{Broader Impact and Limitation}
Due to the hierarchical framework of our proposed F2L, each sub-region acts like an independent FL process. Therefore, our F2L is integrable with other current methods, which means that we can apply varying FL techniques (e.g., FedProx, FedDyne, FedNova, HCFL \cite{2022-FL-HCFL}) into distinct sub-regions to enhance the overall F2L framework. Therefore, architecture search (e.g., which FL technique is suitable for distinct sub FL region) for the entire hierarchical network is essential for our proposed framework, which is the potential research for the future work. Moreover, the hierarchical framework remains unearthed. Therefore, a potentially huge amount of research directions is expected to be investigated (e.g., resource allocation \cite{2022-FL-ResourceAllocation-Compression, 2022-FL-StragglingPrivacy, 2021-FL-CodedFL,2021-FL-LargeScale}, and task offloading in hierarchical FL \cite{2021-FL-EnergyEfficiency}). However, our LKD technique still lacks of understanding about the teachers' models (e.g., how classification boundaries on each layer impact on the entire teachers' performance). By investigating in explainable AI, along with layer-wise performance, we can enhance the LKD, along with reducing the unlabeled data requirements for the distillation process in the future work.
\section{Conclusion}
In this research, we have proposed an FL technique that enables knowledge distillation to extract the-good-feature-only from clients to the global model. Our model is capable of tackling the FL's heterogeneity efficiently. Moreover, experimental evaluations have revealed that our F2L model outperforms all of the state-of-the-art FL baselines in recent years. 
\bibliography{iclr2023_conference.bib}
\bibliographystyle{iclr2023_conference}
\clearpage
\appendix
\section{Background}
\subsection{Stochastic Gradient Descent and Federated Learning}
\label{sec:Background:SGDandFL}
Consider an optimization on a Deep Neural Network (DNN) with the SGD algorithm and a set of parameters $\boldsymbol{\omega} = \{\omega_1, \omega_2, \dots, \omega_p\}$ with $p$ being the size of the DNN's model. SGD \cite{1951-SGD-Fundamental} usually uses a mini-batch gradient \cite{2020-SGD-MinibatchvsLocal} $\widetilde{g}(\boldsymbol{\omega})=-\nabla_{\boldsymbol{\omega}}g_b(\boldsymbol{\omega})$ instead of a full-batch gradient $g(\boldsymbol{\omega})=-\nabla_{\boldsymbol{\omega}}f(\boldsymbol{\omega})$. The subscript $b$ denotes the mini-batch index set, which is drawn from $B$ batches randomly. Therefore, in a mini-batch SGD, the full-batch dataset $\mathcal{D}$ is divided into $B$ mini batches: $\mathcal{D} = \{\mathcal{D}_1, \mathcal{D}_2, \dots, \mathcal{D}_B\}$. Thus, the mini-batch SGD satisfies: 
\begin{equation}\label{eq:SGD-minibatch-fullbatch}
\begin{split}
   g(\boldsymbol{\omega}) = \frac{1}{B}\sum^ {B}_{b=1}{\widetilde{g}(\boldsymbol{\omega})}.
\end{split}
\end{equation}
FL utilizes the concept of SGD. In terms of implicit optimization for FL, clients collect data in the areas which are under their supervision. Thus, data, collected by FL clients from each communication round, can act as a mini-batch of the SGD (the whole data in the FL system counts as the full-batch data) and contribute to the local training process on distributed clients. On completion of the local training at clients, the clients then send their local trained loss functions to the server for aggregation process \cite{2017-FL-FederatedLearning}. The original idea of implicit optimization in FL is similar to the relationship between the full-batch and mini-batch gradients in \eqref{eq:SGD-minibatch-fullbatch}, which is as follows:
\begin{equation}\label{eq:FL-Aggregation-Loss}
\begin{split}
   f(\boldsymbol{\omega}) = \frac{1}{N}\sum^ {N}_{n=1}{f_n(\boldsymbol{\omega})},
\end{split}
\end{equation}
where $f(\boldsymbol{\omega})$ is the aggregated loss function at the server, and $f_n(\boldsymbol{\omega})$ is the resulting loss at the client $n$ in the system with $N$ clients. In order to simplify the FL process, the authors in \cite{2017-FL-FederatedLearning} proposed the surrogate function, namely FedAvg, in which the generality of the FL process is preserved: 
\begin{equation}\label{eq:FL-Aggregation-Weight}
\begin{split}
   \widetilde{\boldsymbol{\omega}} = \frac{1}{N}\sum^ {N}_{n=1}{\boldsymbol{\omega}_n}.
\end{split}
\end{equation}
We have $\widetilde{\boldsymbol{\omega}}$ to be the global parameter set that is aggregated from the set of $N$ clients in the system. Due to the aforementioned similarities between SGD and FL, FL possesses all characteristics of the SGD.
 
\subsection{Knowledge Distillation}
\label{appendix:knowledge-distillation}
Knowledge Distillation (KD) \cite{2015-DL-KnowledgeDistillation} employs a technique to transfer the learned knowledge from a pre-trained teacher model to another model with less or similar complexity (student model). The model transfer process is implemented in two steps. In the first step, a surrogate output probability function, namely a temperature-softmax function, is utilized. By adding a temperature scaling variable $T$, the conventional softmax function then becomes:
\begin{equation}\label{KD-studentOutput}
\begin{gathered} 
\hat{p}(l|\boldsymbol{X}, \boldsymbol{\omega}, T) = \frac{\text{exp}(z^l(\boldsymbol{X})/T)}{\sum^C_{j=1}\text{exp}(z^j(\boldsymbol{X})/T)},  
\end{gathered}
\end{equation}
where $z^j$ is the output set corresponding to class $j$ of the given DNN with batch of input data $\boldsymbol{X}$. The subscript $l$ denotes the index of softmax output which corresponds to the prediction on class $l$ of the DNN. The intention of adding the variable $T$ is to adjust the slope of the softmax function in the classifier as shown in \eqref{KD-studentOutput}. As we can see, when we increase the value of $T$, the slope of the softmax function will decrease significantly. With large temperature scale values, over the same output range of data, the range of values represented by input $z^j$ is larger. Then, the output value tuple created by the teacher and student carries considerably more information. Therefore, the learning process between teacher and student is more effective.

In the second step, to help implement the transfer of knowledge from teacher to student, the authors in \cite{2015-DL-KnowledgeDistillation} presented a new loss function, called distillation loss function (also known as soft-loss function). This loss function comprises two terms: the intrinsic entropy function of the teacher and the cross-entropy function between the teacher and the student's outputs, which can be expressed as follows: 
\begin{align}\label{KD-SoftLoss} 
\mathcal{L}_\text{KD} &= H(p(\boldsymbol{X}),q(\boldsymbol{X})) - H(p(\boldsymbol{X})) 
= \sum_{x_i \in \boldsymbol{X}}{p(x_i) \log{p(x_i)}} - \sum_{x_i \in \boldsymbol{X}}{p(x_i) \log{q(x_i)}}.
\end{align}
The purpose of this function is to compare the output distribution between teacher $p(\boldsymbol{X})$ and student model $q(\boldsymbol{X})$. In terms of information theory, this measurement shows the under-performance of the distribution set created by student, when the output distribution set of the teacher is taken as the sample distribution set, with precision of $100\%$. By minimizing this function, we reduce the functional difference between the two deep networks. As a result, the student model tends to make it's output distribution become more identical to the teacher model's behavior.

Moreover, to restrain the student model from learning the teacher's false result, the authors in \cite{2015-DL-KnowledgeDistillation} proposed the hard-loss formula:
\begin{align}\label{KD-HardLoss} 
\mathcal{L}_\text{CE} &= H(P, \hat{P}) = - \sum_{x_i \in \boldsymbol{X}}{p(x_i) \log{\hat{p}(x_i)}}.  
\end{align}
The hard-loss formula acts as a backing force with two main objectives. First of all, it ensures that the joint distillation function always follows the right optimizing track. Secondly, due to the highly complicated function made by the soft-loss function, the hard-loss force aids the model to escape the local minima whenever the model gets trapped. In this way, student's training performance improves significantly.

\clearpage
\section{Issues about SGD and Federated Learning}
\label{appendix:SGDissues}
As mentioned in Appendix~\ref{sec:Background:SGDandFL}, FL and SGD have the same attributes. Occasionally, SGD is an efficient method for DL. For a large dataset, SGD can converge faster as it causes updates to the parameters more frequently. Moreover, the steps taken towards the minima of the loss function have oscillations \cite{2017-DL-DDN-via-Information} that can help to get out of the local minima of the loss function \cite{2015-SGD-AddingNoises2Gradient,2020-SGD-Noise-GeneralizationBenefit, 2019-DL-SharpFlatMinima}. Therefore, FL can operate well under the ideal conditions. However, when the problems, such as heterogeneity on data and high complexity on training models come about, both SGD and FL face many obstacles.
\subsection{The generalization gap in SGD and FL}
\label{subsec:SGDissues-generalizationGap}

\begin{figure}[h]
\centerline{\includegraphics[width=0.5\linewidth]{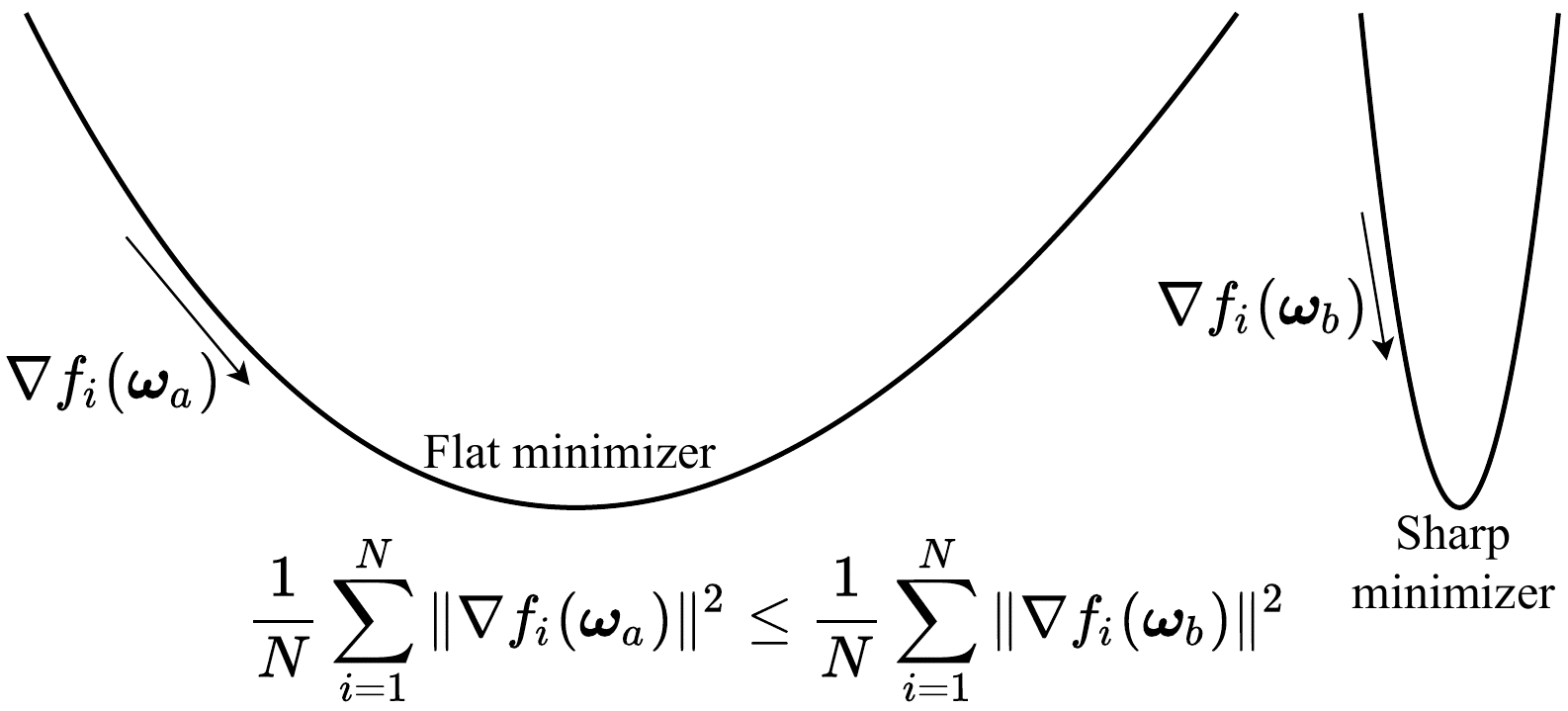}}
\caption{Illustration of flat and sharp minima.}
\label{fig:MinimaShape}
\end{figure}

By measuring the distance between initial model and trained one, the research in \cite{2018-SGD-BadGradient} manifests the proof of low distance traveled by the high complexity model using SGD. The authors observed that Vanilla SGD suffers from high overfitting compared to other training optimizing methods, such as data augmentation, $L2$-regularization, and momentum. Moreover, the model's travelled distance which is trained by Vanilla SGD is $4$ to $5$ times lower than that is trained by the other methods. The research in \cite{2017-SGD-LargeBatchTraining-GeneralizationGap} shows that the lack of generalization ability is because large-batch methods tend to converge to sharp minimizers of the training function. These minimizers are characterized by a significant number of large positive eigenvalues in Hessian matrix of the loss function $f(\boldsymbol{\omega})$, and tend to generalize less effectively. When considering each distributed node in the FL system as a mini-batch of SGD, we can imagine the current FL system as a large-batch for SGD, which can easily be trapped in the sharp minimum. For a more mathematical analysis, we consider the gradient dissimilarity inequality for the aggregated global model $\boldsymbol{\omega}$, there exist constants $G\geq0$ and $U\geq1$ that attain:
\begin{align}\label{eq:Gradient-Dissimilarity} 
    \frac{1}{N}\sum^N_{n=1}{\Vert \nabla f_n(\boldsymbol{\omega}) \Vert^2} \leq G^2 + U^2 \Vert \nabla f(\boldsymbol{\omega}) \Vert^2, ~\forall \boldsymbol{\omega}.
\end{align}
Here, we follow the heterogeneity assumption that $G$ is defined with gradient variance boundary (on client $n$) $\sigma^2_n$ as in \cite{2019-FL-CompressGradientDifferences}. Along with the condition that the problem $f_i(\boldsymbol{\omega})$ is $\mu$-smooth, the assumption can be relaxed as in \cite{2020-FL-Scaffold}: 
\begin{align}\label{eq:Gradient-Dissimilarity-Smooth} 
    \frac{1}{N}\sum^N_{i=1}{\Vert \nabla f_i(\boldsymbol{\omega}) \Vert^2} - G^2 
    \approx \underbrace{\frac{1}{N}\sum^N_{i=1}{\Vert \nabla f_i(\boldsymbol{\omega}) \Vert^2}}_{\mathbb{A}} - \underbrace{\frac{2}{N}\sum\limits^N_{i=1}{\Vert\nabla f_i(\boldsymbol{\omega}^*) \Vert^2}}_{\mathbb{B}}
    \leq 2\mu B^2 (f(\boldsymbol{\omega})-f^*),
\end{align}
 where $f^*$ denotes the optimality. From this inequality, we can see that the distance between the current state $f(\boldsymbol{\omega})$ and the global optima state $f^*$ is lower-bounded by $\mathbb{A}+\mathbb{B}$. As we can see from \eqref{eq:Gradient-Dissimilarity-Smooth}, the first term $\mathbb{A} = \frac{1}{N}\sum^N_{i=1}{\Vert \nabla f_i(\boldsymbol{\omega}) \Vert^2}$ reveals the average gradient norm at the model state $\boldsymbol{\omega}$. When this term of value is large, the current state is assumed to have a high chance to be far from the true optimal state $f^*$ (due to the term $\mathbb{B}$ being fixed). Simultaneously, the current state $\boldsymbol{\omega}$ is more likely to be a sharp minimizer because of a larger $\mathbb{A}$ value. The illustrative explanation about sharp and flat minimizers is presented in Fig.~\ref{fig:MinimaShape}.  
\subsection{Client-drift on FL}
\begin{figure}[!h]
\centerline{\includegraphics[width=0.5\linewidth]{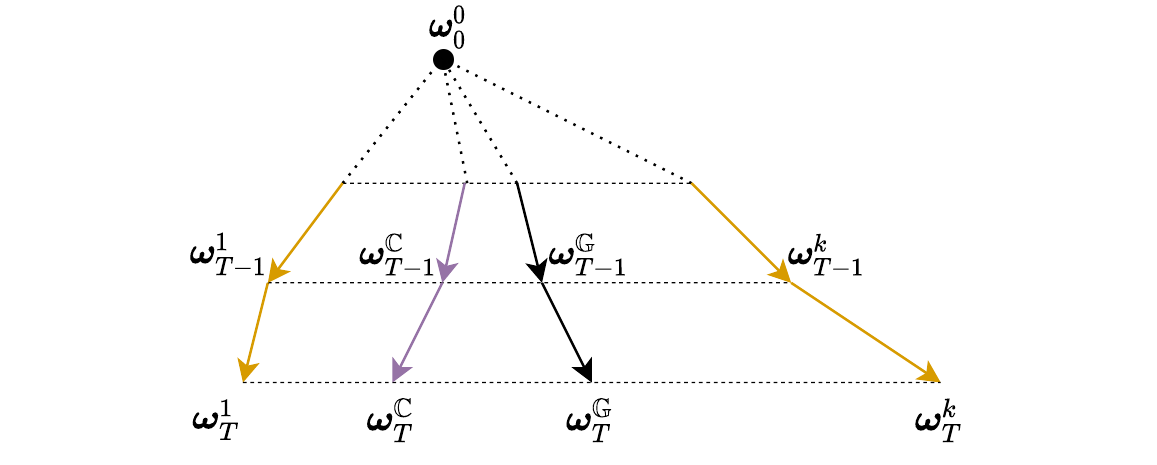}}
\caption{Illustration of weight divergence in non-IID data.}
\label{fig:WeightDivergence}
\end{figure}

The authors in \cite{2018-FL-NonIIDAnalysis} introduced a term named \say{weight divergence}, which can be computed as $\Vert \boldsymbol{\omega}^\mathbb{G} - \boldsymbol{\omega}^\mathbb{C} \Vert$.
This weight divergence is the total distance between two models: practical aggregated global model of the FL system $\boldsymbol{\omega}^\mathbb{G}$ and ideal model when trained with the data collected and shuffled one place $\boldsymbol{\omega}^\mathbb{C}$. For a mathematical explanation, weight divergence at epoch $E$ of communication round $m$ can be demonstrated as follows: 
\begin{align}\label{eq:Weight-Divergence}
    &\Vert \boldsymbol{\omega}^\mathbb{G}_{mE} - \boldsymbol{\omega}^\mathbb{C}_{mE} \Vert \leq \sum^N_{n=1}  \frac{s^n}{\sum^N_{n=1}s^n}(a^n)^E \Vert \boldsymbol{\omega}^\mathbb{G}_{(m-1)E} - \boldsymbol{\omega}^\mathbb{C}_{(m-1)E} \Vert \notag\\
    &+\eta\sum^N_{n=1}\frac{s^n}{\sum^N_{n=1}s^n}\sum^C_{c=1}\Vert p^n(y=c)-p(y=c) \Vert 
    \times \sum^{E-1}_{e=0}(a^n)^e g_\text{max}(\boldsymbol{\omega}^\mathbb{C}_{mE-1-n}), 
\end{align}
where $\eta$ is the learning rate, $s^n$ is the number of samples on client $n$ in a set of $N$ clients, $p^n(y=c)$ is the probability density of the class $c$ on client $n$, and $p(y=c)$ is the global model's probability density. $g_\text{max}(\boldsymbol{\omega}^t_{mE-1-n})$ is the max gradient over $C$ classes, and $a^n$ is the average smooth coefficient over $C$ classes which is defined as: $1+\eta\sum^C_{c=1}p^n(y=c)\mu_{c}$ with assumption that the problem $f_i(\boldsymbol{\omega})$ is $\mu_{c}$-smooth on class $c$. As mentioned in \cite{2018-FL-NonIIDAnalysis}, we have three following observations: 
\begin{itemize}
    \item The weight divergence is affected by two aspects: the weight divergence value from the last global aggregation, and the relative entropy between data distribution on client $n$ and the actual distribution on the whole population. 
    \item The term $\sum^N_{n=1}  \frac{s^n}{\sum^N_{n=1}s^n}(a^n)^E$ acts as an amplifier. Regardless of the data property, the weight divergence still occurs to the aforementioned amplifier term. Therefore, the FL system always suffers the reduction in accuracy even if the data is IID.   
    \item When all clients start from the same initialization, the probability distance $\Vert p^n(y=c)-p(y=c) \Vert$ is the matter of divergence. The illustration of the probability distance is described in Fig.~\ref{fig:WeightDivergence} 
\end{itemize}

\section{System Model}
\label{sec:system-model}
 As shown in Fig.~\ref{fig:hierarchicalFL}, we first compartmentalize the network into $R$, where each region contains one access point (AP) acting as a local server (denoted by $r \in \{1,2,\dots, R\}$) and $N$ clients (denoted by $n \in \{1,2,\dots,N\}$) are randomly sampled from the FL system. For each communication round of FL, those APs execute the FedAvg, which aggregates the model parameters from local predictors in their own regions, resulting in the regional aggregated model $\boldsymbol{\omega}^r$. The global server receives the aggregated model parameters after a certain number of communication rounds. Each updated model, thus, acts as a teacher to share knowledge to the global deep model $\boldsymbol{\omega}^g$. The data pool $\mathcal{S}=(\boldsymbol{X}, \boldsymbol{Y}) = \{(x_i, y_i)\}$ in which $i \in \{1,2,\dots,S\}$ is stored in the global server for the multi-teacher distillation process. Here, $y_i \in \{1,2,\dots,C\}$ is the truth label of the preset dataset $\mathcal{S}$. We define $l$ as the output index of the deep model and $c \in \{1,2,\dots,C\}$ as the predicted label index of the data $\boldsymbol{X}$ when it is processed by the deep model.
 
\section{F2L algorithm details}

\begin{algorithm}[h]
    \caption{Label-driven Joint KD}
    \label{alg:LD-KD}
\begin{algorithmic}
    \State {\bfseries Input:} $\boldsymbol{\omega} = (\boldsymbol{\omega}^1,\boldsymbol{\omega}^2,\dots,\boldsymbol{\omega}^R), \boldsymbol{\omega}^g$, $(\boldsymbol{X}, \boldsymbol{Y})$
    \State initialize $\mathcal{L}_\text{LKD} = 0$
    \For{each epoch e = 1,2,\dots, E}
        \For{each region r = 1,2,\dots, R}
            \State $\boldsymbol{\beta}_r\gets\text{C-Reliability}(\boldsymbol{\omega}^r,\boldsymbol{X})$ in Algorithm~\ref{alg:C-Reliability}
            \State $\boldsymbol{X}^r_\text{alg}, \boldsymbol{Y}^r_\text{alg} \gets\text{L-SampleAlign}(\boldsymbol{X}, r, \boldsymbol{\omega}^r)$ in Algorithm~\ref{alg:L-SampleAlign}
            \State $\mathcal{L}_\text{LKD} \gets \text{L-KD}(\boldsymbol{\omega}^r,\boldsymbol{X}^r_\textit{alg}, \boldsymbol{Y}^r_\textit{alg},\boldsymbol{\beta}_r)$ in Algorithm~\ref{alg:L-KD}
        \EndFor
        \State $\boldsymbol{X}^g_\text{alg}, \boldsymbol{Y}^g_\text{alg} \gets\text{L-SampleAlign}(\boldsymbol{X}, r, \boldsymbol{\omega}^g)$
        \State $\mathcal{L}_\text{LKD} \gets \mathcal{L}_\text{LKD} + \text{G-update-KD}(\boldsymbol{\omega},\boldsymbol{X}^g_\text{alg}, \boldsymbol{Y}^g_\text{alg})$
        \State $\mathcal{L}_\text{LKD} \gets \mathcal{L}_\text{LKD} + \mathcal{L}_\text{CE}(\boldsymbol{\omega}^g, \boldsymbol{X}, \boldsymbol{Y})$
        \State Update the global parameters:
        \State $\boldsymbol{\omega^g} \leftarrow \boldsymbol{\omega^g} - \eta \nabla \mathcal{L}_\text{LKD}$ with $\eta$ is the learning rate
    \EndFor
\end{algorithmic}
\end{algorithm}

\begin{algorithm}[h]
    \caption{L-SampleAlign}
    \label{alg:L-SampleAlign}
\begin{algorithmic}
    \State {\bfseries Input:} $\boldsymbol{X}$, $r$, $\boldsymbol{\omega}^r$ 
    \For{$x_i$ in $\boldsymbol{X}$}
        \State $c \gets \text{Predict label on model } \boldsymbol{\omega}^r$
        \State $\boldsymbol{X}^r_\textit{alg}[c], \boldsymbol{Y}^r_\textit{alg}[c]  \gets x_i, y_i$
    \EndFor
\end{algorithmic}
\end{algorithm}
\clearpage
\begin{algorithm}[h]
    \caption{L-KD}
    \label{alg:L-KD}
\begin{algorithmic}
    \State {\bfseries Input:} $\boldsymbol{\omega}^r,\boldsymbol{X}^r_\textit{alg}, \boldsymbol{Y}^r_\textit{alg},\boldsymbol{\beta}$
    \State Initialize: $\mathcal{L}_r = 0$
    \For{$c$ in $(1,2,\dots,C)$)}
        \For{$x_i$ in $\boldsymbol{X}^r_\text{alg}[c]$}
            \State $\boldsymbol{p}_i \gets \text{Predict logit output on model }\boldsymbol{\omega}^r$
            \State $\boldsymbol{q}_i \gets \text{Predict logit output on model }\boldsymbol{\omega}^g$
            \For{$l$ in $(1,2,\dots, C)$}
              \State $\textit{KL}_l = p_i^l \times (\log{p_i^l} - \log{q_i^l})$
              \State $\mathcal{L}_r \gets \mathcal{L}_r + \textit{KL}_l$
            \EndFor
        \EndFor
    \EndFor
\end{algorithmic}
\end{algorithm}
\begin{algorithm}[h]
    \caption{G-Update-KD}
    \label{alg:G-Update-KD}
\begin{algorithmic}
    \State {\bfseries Input:} $\boldsymbol{\omega},\boldsymbol{X}, \boldsymbol{Y}$
    \For{$x_i$ in $\boldsymbol{X}^g_\text{alg}$}
    \For{$l$ in $(1,2,\dots, C)$}
        \State ${p}_i^l = {q}_i^l$
        \State $\boldsymbol{q}_i \gets \text{Predict logit output on model }\boldsymbol{\omega}^g$
        \State $\textit{KL}_\text{upd} = p_i^l \times (\log{p_i^l} - \log{q_i^l})$
        \State $\mathcal{L} = \mathcal{L} + \textit{KL}_\text{upd}$
    \EndFor
    \EndFor  
\end{algorithmic}
\end{algorithm}
\begin{algorithm}[h]
    \caption{C-Reliability}
    \label{alg:C-Reliability}
\begin{algorithmic}
    \State {\bfseries Input:} $\boldsymbol{\omega}^r,\boldsymbol{X}$
    \State Initialize: 
    \For{$r$ in $1,2,\dots,R$}
        \State $S_r^l = 0$
    \EndFor
    \For{$r$ in $1,2,\dots,R$}
        \State $f_\text{AUC}^r =  \text{evaluate AUC on model }\boldsymbol{\omega}^r$ 
        \For{$l$ in $1,2,\dots,C$}
            \State $S_r^l \gets S_r^l + \exp(f_\text{AUC}^{l,r})$ with $f_\text{AUC}^{l,r} \in f_\text{AUC}^r$
        \EndFor
    \EndFor
    \For{$l$ in $1,2,\dots,C$}
        \State $\beta^c_r \gets \exp(f^{l,r}_\text{AUC})/S_r^l$
    \EndFor
\end{algorithmic}
\end{algorithm}
\clearpage 
\section{Experiment Setup}
\label{sec:V-A-ExperimentSetup}
\textbf{Dataset.} We evaluate all algorithms on non-IID data partitioning with Dirichlet random distribution function \cite{2021-FL-ExperimentalStudy}. We use benchmark datasets with the same train/test splits as in \cite{2017-FL-FederatedLearning}. We use four multi-class categorisation benchmark datasets in our evaluations, including \textbf{1)} MNIST \cite{2010-DL-MNIST}, \textbf{2)} EMNIST \cite{2017-DL-EMNIST}, \textbf{3)} CIFAR-10 \cite{2010-DL-CIFAR}, \textbf{4)} CIFAR-100 \cite{2010-DL-CIFAR}, \textbf{5)} CINIC-10 \cite{2020-DL-CINIC10}, and \textbf{6)} CelebA gender classification \cite{2020-DL-CelebA} . The evaluation settings for simulations are listed in Appendix~\ref{appendix:data-parameter-settings}.

\textbf{Settings.}
We apply two DNNs for our experimental evaluations, including LeNet-5 \cite{1989-DL-LeNet5} for MNIST and EMNIST datasets, and ResNet-$18$ \cite{2015-DL-ResNet} for CIFAR-$10$ and CIFAR-$100$. It is worth noting that we do not apply pre-trained models on both LeNet-$5$ and ResNet-$18$ because the pre-trained models can help FL approach the global minima easily, which can reduce the generalization of the evaluation. The detailed settings is defined in Appendix~\ref{appendix:data-parameter-settings} 

\textbf{Baselines.} We compare the performance on non-IID dataset with four baselines, including FedProx, FedDistill \cite{2021-FL-FedDistill}, FedGen \cite{2021-FL-FedGen}, and FedAvg. 

\textbf{Evaluation metric.} We adopt top-1 accuracy on every experiments to evaluate the performance of our F2L method and the other FL baselines. Further, each evaluation is an average of five results. 

\textbf{Settings for Scalability test.} We use CINIC-10 dataset, where the data consists of $270.000$ images ($90.000$ images in each train, validation, test set). We use the train set for the FL training, validation set for the accuracy evaluation. To make the dataset for the additional clients, we use the test set. The more detailed of the settings is demonstrated in Table~\ref{tab:scalability-test}. As described in Table~\ref{tab:scalability-test}, the data on every client is sampled with non-IID coeff $\alpha = 0.1$ (on both old and new regions). To make the test between FedAvg and F2L fair, we use the same number of users for added regions (i.e., total added clients is $100$). 

\begin{table}[!h]
\renewcommand{\arraystretch}{1.25}
\caption{The data settings used for scalability test.\\}
\centering 
\small\addtolength{\tabcolsep}{-3pt}
\begin{tabular}{| l | c | c | c |} 
\hline
\multicolumn{4}{c}{\textbf{Data Setting}} \\
\hline
&Old & Added & Evaluation \\ [0.5ex] 
&Regions & Regions & dataset \\ [0.5ex] 
\hline  
Data size          & $90000$  & $90000$ & $90000$ \\ 
non-IID coeff $\alpha$ & $1$ & $0.1$ & N/A       \\  
\hline
\multicolumn{4}{c}{\textbf{Vanilla FL system}}      \\
\hline
Clients per region  & $100$ & $100$ & $1$       \\  
Samples per client  & $5000$  & $5000$ & $10000$    \\  
\hline
\multicolumn{4}{c}{\textbf{F2L system}}      \\
\hline
Clients per region  & $33,33,34$ & $33,33,34$ & $1$       \\  
Samples per client  & $5000$  & $5000$ & $10000$    \\  
Samples on server  & $9000$  &  &     \\  
\hline
\end{tabular}
\label{tab:scalability-test} 
\end{table}

\clearpage
\section{Proof on Random Sampling effects to Data Distribution}
\label{appendix:proof-on-sampling-and-data-distribution}
\begin{assumption}
   We consider the class-likelihood of an data as the classical Gaussian Mixture Model (GMM). Therefore, the feature distribution on data can be represented as the Dirichlet Process GMM (DPGMM) \cite{2018-MF-DirichletProcessGaussianMixtureModel}.
\begin{align}
    p\left(x \vert (\mu_c, \Sigma_c, \pi_c)^C_{c=1}\right)
    = 
    \sum^C_{c=1} \pi_c ~ \mathcal{N}(\mathbf{x}; \mu_c, \Sigma_c),
\end{align}
where $\pi_c$ is the Dirichlet allocation weight on class $c$.
\label{assumption:data-DPGMM}
\end{assumption}
\begin{lemma}
    The data distribution on regional FL is a combination of client set, and can be represented as a DPGMM. 
\begin{align}
    p\left(x \vert (\mu_c, \Sigma_c, \pi_c)^C_{c=1}\right)
    = 
    \sum^{C}_{c=1} \alpha_{c} ~ \mathcal{N}(\mathbf{x}; \mu_c, \Sigma_c),    
\end{align}
where $\alpha_{c}$ is the Dirichlet allocation weight on class $c$ of data point $j$ over total $S_c^r$ data sampled from region $r$.
\label{lemma:regional-DPGMM}
\end{lemma}
\textit{Proof:} The proof is demonstrate in Appendix~\ref{appendix:regional-DPGMM}.

To theoretically analyze the random sampling effects to regional data distribution, we need to prove that $\pi^{r}_{c} \neq \pi^{r}_{c'}$ and $\pi^{r}_{c} \neq \pi^{r'}_{c}$. To this end, we consider covariance $\Cov(\pi^{r}_{c}, \pi^{r'}_{c})$ and $\Cov(\pi^{r}_{c}, \pi^{r}_{c'})$. Applying covariance equation for Dirichlet distribution \cite{2003-MF-InformationTheory-LearningAlgo}, we have:
\begin{align}
    \Cov(\pi^{r}_{c}, \pi^{r}_{c'}) = - \frac{\nu^{r}_{c} \nu^{r}_{c'}}{\Bar{\nu}^2(\Bar{\nu}+1)}, 
\end{align}
\begin{align}
    \Cov(\pi^{r}_{c}, \pi^{r'}_{c}) = - \frac{\nu^{r}_{c} \nu^{r'}_{c}}{\Bar{\nu}^2(\Bar{\nu}+1)}. 
\end{align}
As we can see from the two mentioned equations, the covariances $\Cov(\pi^{r}_{c}, \pi^{r'}_{c})$ and $\Cov(\pi^{r}_{c}, \pi^{r}_{c'})$ is always lower than $0$. Furthermore, when a system is suffered more seriously from non-IID, the concentration parameter (i.e., $\nu^{r}_{c}$) becomes higher \cite{2019-MeasuringEffects-FL}. This phenomenon makes the random sampling capable of trigger high class-wise bias on regional FL. Therefore, the F2L can be more effective on heterogeneous network by applying LKD (which is more briefly described in \ref{sec:IV-D-LabelDrivenKD}).

\clearpage
\section{Label Driven Knowledge Distillation Decomposition}
\label{appendix:labeldriven-knowledgedistillation}
Recall the KL divergence function in \cite{2005-IF-KLdivergence}, the KL divergence can be expressed as the average distance between teacher and student's probability among all input value $x_i \in \boldsymbol{X}$. Combining the aforementioned statement with \eqref{LKD-surrogateOutput}, we can decompose the KL divergence function into multiple sub-components as: 
\begin{align}
    \mathcal{L}_r^\textit{KL} 
    &= \sum^C_{l=1} \sum^{S}_{i=1} \hat{p}^r(l|x_i,\boldsymbol{\omega}^r,T,c)\log{\frac{\hat{p}^r(l|x_i,\boldsymbol{\omega}^r,T,c)}{\hat{p}^g(l|x_i,\boldsymbol{\omega}^g,T,c)}} \notag\\
    &= \sum^C_{l=1} \sum^C_{c=1} \sum^{S_c^r}_{i=1} \hat{p}^r(l|x_i,\boldsymbol{\omega}^r,T,c)\log{\frac{\hat{p}^r(l|x_i,\boldsymbol{\omega}^r,T,c)}{\hat{p}^g(l|x_i,\boldsymbol{\omega}^g,T,c)}} 
    = \sum^C_{c=1} D_\textit{KL}^c(\hat{p}^r||\hat{p}^g).
\label{eq:general-mtkd-decomposed}
\end{align}
Here, $S$ is the number of samples of the fixed dataset $\mathcal{S}$ on the server. $(\boldsymbol{X}^r_\text{alg}, \boldsymbol{Y}^r_\text{alg})$ is the dataset which is pseudo labeled and aligned by regional model $r$, and we have $(\boldsymbol{X}^r_\text{alg}[c], \boldsymbol{Y}^r_\text{alg}[c])$ as the set of data with size of $S_c^r$ labeled by the model $r$ as $c$. 

Although the same preset dataset is utilized on every teacher model, the different pseudo labeling judgments from different teachers lead to the different dataset tuples. The process of identifying $S_c^r$ is demonstrated in Algorithm~\ref{alg:L-SampleAlign}. Because the regional models label on the same dataset $S$, we have $\sum_{c=1}^C S^r_c = S$ for all regional models. $D_\textit{KL}^c(\hat{p}^r||\hat{p}^g)$ is the $c$ label-driven KL divergence between model $r$ and model $g$. 

By dividing the dataset by class (regarding the teacher's label-driven prediction), we have the teacher's prediction as the pseudo label. Therefore, we can leverage the unlabeled dataset to train the LKD at the server. Moreover, with the decomposed equation~\eqref{eq:general-mtkd-decomposed}, we can scale the point-wise distribution distance between two model $r$, and $g$. Therefore, the scale can be leveraged to judge the teacher model's label driven performance, and then, the good knowledge can be distilled. The surrogate MTKD function, so called LKD is demonstrated as follows: 
\begin{align}
    \textbf{P3}:\min \mathcal{L}_m^\textit{KL} 
    &= \sum^R_{r=1} \sum^C_{c=1} \beta^c_r \sum^{S_c^r}_{i=1} \sum^C_{l=1} \hat{p}^r(l|x_i,\boldsymbol{\omega}^r,T,c)\log{\frac{\hat{p}^r(l|x_i,\boldsymbol{\omega}^r,T,c)}{\hat{p}^g(l|x_i,\boldsymbol{\omega}^g,T,c)}} \notag\\
    &= \sum^R_{r=1} \sum^C_{c=1} \beta^c_r D_\textit{KL}^c(\hat{p}^r||\hat{p}^g).
\label{eq:general-lkd}
\end{align}

\section{Proof on Theorem~\ref{theorem:labeldriven-knowledgedistillation-analysis}}
\label{appendix:labeldriven-knowledgedistillation-analysis}

We consider the sum
\begin{align}
    S = \sum^{n}_{j=1} \sum^{n}_{k=1} (e^{\tau^c_j}-e^{\tau^c_k})(\sigma^2_{j,c} - \sigma^2_{k,c}).
\end{align}
We have $\sigma^2_{r,c}$ sequence is non-increasing. Furthermore, the $e^{\tau^c_r}$ sequence is non-decreasing due to the \ref{appendix:relationship-accuracy-variance} and Jensen's inequality \cite{2010-MF-Probability}. Therefore $(e^{\tau^c_j}-e^{\tau^c_k})(\sigma^2_{j,c} - \sigma^2_{k,c}) \leq 0 \quad \forall j,k$, and thus $S \leq 0$. Thus, we have:
\begin{align}
    & \frac{2R}{\sum_{r=1}^{R} e^{\tau^c_r}} \sum^{R}_{r=1} e^{\tau^c_j} \sigma^2_{r,c} 
    - \frac{2}{\sum_{r=1}^{R} e^{\tau^c_r}} \sum^{R}_{r=1} e^{\tau^c_r} \sum^{R}_{k=1} \sigma^2_{r,c} \leq 0, 
\end{align}
which also means:
\begin{align}
    \frac{2R}{\sum_{r=1}^{R} e^{\tau^c_r}} \sum^{R}_{r=1} e^{\tau^c_r} \sigma^2_{r,c} \leq \frac{2}{\sum_{r=1}^{R} e^{\tau^c_r}} \sum^{R}_{r=1} e^{\tau^c_r} \sum^{R}_{r=1} \sigma^2_{r,c}. 
\label{eq:LKD-2-MTKD-1-variance}
\end{align}

From \eqref{eq:LKD-2-MTKD-1-variance}, we can have the following deduction to the relationship between student models distilled by LKD and MTKD:
\begin{align}
    \sigma^{*2}_{\rm{LKD},g,c} = \frac{1}{\sum_{r=1}^{R} e^{\tau^c_r}} \sum^{R}_{r=1} e^{\tau^c_r} \sigma^2_{r,c} \leq (\frac{\sum^{n}_{r=1} e^{\tau^c_r}}{\sum_{r=1}^{R} e^{\tau^c_r}})(\frac{1}{R} \sum^{R}_{r=1} \sigma^2_{r,c}) =
    \frac{1}{R} \sum^{R}_{r=1} \sigma^2_{r,c} \overset{(a)}{=} 
    \sigma^{*2}_{\rm{MTKD,g,c}}.
\end{align}
We can prove $(a)$ based on Lemma~\ref{lemma:optimal-LKD} by setting the LKD allocation weight set $\beta^c_1 = \beta^c_2 = \dots = \beta^c_R = 1$. 

\section{Proof on Theorem~\ref{theorem:labeldriven-knowledgedistillation-analysis-mean}}
\label{appendix:labeldriven-knowledgedistillation-analysis-mean}

We consider the sum
\begin{align}
    S = \sum^{n}_{j=1} \sum^{n}_{k=1} (e^{\tau^c_j}-e^{\tau^c_k})(|\mu_{j,c} - \Bar{\mu}_c|-|\mu_{k,c} - \Bar{\mu}_c|).
\end{align}
We have $(|\mu_{j,c} - \Bar{\mu}_c|-|\mu_{k,c} - \Bar{\mu}_c|)$ sequence is non-decreasing. Furthermore, the $e^{\tau^c_r}$ sequence is non-decreasing due to the \ref{appendix:relationship-accuracy-variance} and Jensen's inequality \cite{2010-MF-Probability}. Therefore $(e^{\tau^c_j}-e^{\tau^c_k})(\sigma^2_{j,c} - \sigma^2_{k,c}) \leq 0 \quad \forall j,k$, $S \leq 0$. Thus, we have:
\begin{align}
    & \frac{2R}{\sum_{r=1}^{R} e^{\tau^c_r}} \sum^{R}_{r=1} e^{\tau^c_j} |\mu_{r,c} - \Bar{\mu}_c| 
    - \frac{2}{\sum_{r=1}^{R} e^{\tau^c_r}} \sum^{R}_{r=1} e^{\tau^c_r} \sum^{R}_{k=1} |\mu_{r,c} - \Bar{\mu}_c| \leq 0, 
\end{align}
which also means:
\begin{align}
    \frac{2R}{\sum_{r=1}^{R} e^{\tau^c_r}} \sum^{R}_{r=1} e^{\tau^c_r} |\mu_{r,c} - \Bar{\mu}_c| \leq \frac{2}{\sum_{r=1}^{R} e^{\tau^c_r}} \sum^{R}_{r=1} e^{\tau^c_r} \sum^{R}_{r=1} |\mu_{r,c} - \Bar{\mu}_c|. 
\label{eq:LKD-2-MTKD-1}
\end{align}

From \eqref{eq:LKD-2-MTKD-1}, we can have the following deduction to the relationship between student models distilled by LKD and MTKD:
\begin{align}
    |\mu^{*}_{\rm{LKD},g,c} - \Bar{\mu}_c| &= \frac{1}{\sum_{r=1}^{R} e^{\tau^c_r}} \sum^{R}_{r=1} e^{\tau^c_r} |\mu_{r,c} - \Bar{\mu}_c| \leq (\frac{\sum^{n}_{r=1} e^{\tau^c_r}}{\sum_{r=1}^{R} e^{\tau^c_r}})(\frac{1}{R} \sum^{R}_{r=1} |\mu_{r,c} - \Bar{\mu}_c| \notag \\
    &=
    \frac{1}{R} \sum^{R}_{r=1} |\mu_{r,c} - \Bar{\mu}_c| 
    \overset{(b)}{=} |\mu^{*}_{\rm{MTKD,g,c}} - \Bar{\mu}_c|.
\end{align}
We can prove $(b)$ based on Lemma~\ref{lemma:optimal-LKD} by setting the LKD allocation weight set $\beta^c_1 = \beta^c_2 = \dots = \beta^c_R = 1$. 

\section{Proof on Lemma~\ref{lemma:optimal-LKD}}
\label{appendix:lemma-1}
We consider the KL-divergence equation: 
\begin{align}
    D_\text{KL} = \sum^{S}_{i = 1}{p^r(x_i) \log\frac{p^r(x_i)}{p^g(x_i)}}.
\end{align}
By taking the assumption that each label-driven posterior distribution follows normal distribution from assumption~\ref{assumption:data-DPGMM}, we have the following: 
\begin{align}
    D_\text{KL} 
    &= \sum^{C}_{c = 1}\sum^{S^c_r}_{i = 1}{p^r(x_i \vert y=c) \log\frac{p^r(x_i \vert y=c)}{p^g(x_i \vert y=c)}} \notag \\
    &= \sum^{C}_{c = 1} \E_{p^r(x \vert y=c)} \Big[\frac{1}{2}\Big(\frac{(x_i-\mu_{g,c})^2}{\sigma^2_{g,c}} - \frac{(x_i-\mu_{r,c})^2}{\sigma_{r,c}^2}\Big) + \ln{\frac{\sigma_{r,c}}{\sigma_{g,c}}}\Big] \notag \\
    &= \sum^{C}_{c = 1} \E_{p^r(x \vert y=c)} \Big[\frac{1}{2}\Big(\frac{x_i^2-2\mu_{g,c} x_i + \mu_{g,c}^2}{\sigma^2_{g,c}} - 1 \Big) + \ln{\frac{\sigma_{r,c}}{\sigma_{g,c}}}\Big].
\label{eq:gaussian-KL-divergence}
\end{align}
Applying additional raw moments of the normal distribution
\begin{align}\label{eq:moment-normal-distribution}
X \sim \mathcal{N}(\mu, \sigma^2) \rightarrow
\begin{cases}
    E[x] &= \mu, \\
    E[x^2] &= \mu^2 + \sigma^2, 
\end{cases}
\end{align}
the KL distance in \eqref{eq:gaussian-KL-divergence} becomes:
\begin{align}
    D_\text{KL} 
    &= \sum^{C}_{c = 1} \E_{p^r(x)} \Big[\frac{1}{2}\Big(\frac{x_i^2-2\mu_{g,c} x_i + \mu_{g,c}^2}{\sigma^2_{g,c}} - 1 \Big) - \ln{\frac{\sigma_{r,c}^2}{\sigma_{g,c}^2}}\Big] \notag \\
    &= \sum^{C}_{c = 1} \Big[\frac{1}{2}\Big(\frac{\mu_{r,c}^2 + \sigma_{r,c}^2 - 2\mu_{g,c} \mu_{r,c} + \mu_{g,c}^2}{\sigma^2_{g,c}} - 1 \Big) - \ln{\frac{\sigma_{r,c}^2}{\sigma_{g,c}^2}}\Big] \notag \\
    &= \frac{1}{2} \sum^{C}_{c = 1} \Big[\frac{(\mu_{r,c} - \mu_{g,c})^2}{\sigma^2_{g,c}} + \frac{\sigma^2_{r,c}}{\sigma^2_{g,c}} - 1  - \ln{\frac{\sigma_{r,c}^2}{\sigma_{g,c}^2}}\Big].
\label{eq:gaussian-KL-divergence-shortened}
\end{align}
Combining \eqref{eq:gaussian-KL-divergence-shortened} and LKD from \eqref{eq:general-lkd}, we have:
\begin{align}
    \mathcal{L}_m^\textit{KL} 
    &= \sum^{C}_{c = 1} \sum^R_{r=1} \beta^c_r \Big[\frac{(\mu_{r,c} - \mu_{g,c})^2}{\sigma^2_{g,c}} + \frac{\sigma^2_{r,c}}{\sigma^2_{g,c}} - \ln{\frac{\sigma_{r,c}^2}{\sigma_{g,c}^2}}\Big] \notag \\
    &= \sum^{C}_{c = 1} \frac{1}{\sum_{r=1}^{R} e^{\tau^c}} \sum^R_{r=1} e^{\tau_r^c}\Big[\frac{(\mu_{r,c} - \mu_{g,c})^2}{\sigma^2_{g,c}} + \frac{\sigma^2_{r,c}}{\sigma^2_{g,c}} - \ln{\frac{\sigma_{r,c}^2}{\sigma_{g,c}^2}}\Big] \\
    &= \sum^{C}_{c = 1} \frac{1}{\sum_{r=1}^{R} e^{\tau^c}} \sum^R_{r=1} e^{\tau_r^c} \frac{(\mu_{r,c} - \mu_{g,c})^2}{\sigma^2_{g,c}}
    + 
    \sum^{C}_{c = 1} \frac{1}{\sum_{r=1}^{R} e^{\tau^c}} \sum^R_{r=1} e^{\tau_r^c} \Big[\frac{\sigma^2_{r,c}}{\sigma^2_{g,c}} - \ln{\frac{\sigma_{r,c}^2}{\sigma_{g,c}^2}}\Big]
\label{eq:gaussian-LKD-1}
\end{align}
Firstly, we take $\sum^{C}_{c = 1} (\sum_{r=1}^{R} e^{\tau^c})^{-1} \sum^R_{r=1} e^{\tau_r^c} (\mu_{r,c} - \mu_{g,c})^2(\sigma^2_{g,c})^{-1}$ into consideration, we have:
\begin{align}
    D^c_{B,\textrm{KL}} = \sum^R_{r=1} e^{\tau_r^c} \frac{(\mu_{r,c} - \mu_{g,c})^2}{\sigma^2_{g,c}}
\end{align}
The optimal state of $\sum^R_{r=1} e^{\tau_r^c} ((\mu_{r,c} - \mu_{g,c})^2 / \sigma^2_{g,c})$ is when the $1^\textrm{st}$ derivative equal to $0$. We have: 
\begin{align}
    \nabla D^c_{B,\textrm{KL}} = 2\sum^R_{r=1} e^{\tau_r^c} \frac{(\mu_{r,c} - \mu_{g,c})}{\sigma^2_{g,c}} = 0,
\end{align}
which also means: 
\begin{align}
    \sum^R_{r=1} e^{\tau_r^c} \mu^{*}_{g,c} = \sum^r_{r=1} e^{\tau_r^c} \mu_{r,c}
    \Leftrightarrow
    \mu^{*}_{g,c} \sum^R_{r=1} e^{\tau_r^c}  = \sum^r_{r=1} e^{\tau_r^c} \mu_{r,c} 
    \Leftrightarrow \mu^{*}_{g,c} = \frac{\sum^r_{r=1} e^{\tau_r^c} \mu_{r,c}}{\sum^R_{r=1} e^{\tau_r^c}}
\label{eq:KL-optimal-state-mean}
\end{align}
Taking $\sum^{C}_{c = 1} (\sum_{r=1}^{R} e^{\tau^c})^{-1} \sum^R_{r=1} e^{\tau_r^c} \Big[(\sigma^2_{r,c} / \sigma^2_{g,c}) - \ln{(\sigma_{r,c}^2 / \sigma_{g,c}^2})\Big]$ is when the derivative of $D^c_{B,\textrm{KL}}$ into consideration, we have: 
\begin{align}
    \mathcal{L}_m^\textit{KL} 
    &= \sum^{C}_{c = 1} \frac{1}{\sum_{r=1}^{R} e^{\tau^c}} \sum^R_{r=1} e^{\tau_r^c}\Big[\frac{\sigma^2_{r,c}}{\sigma^2_{g,c}} - \ln{\frac{\sigma_{r,c}^2}{\sigma_{g,c}^2}}\Big].
\end{align}
We consider the label-driven KL divergence: 
\begin{align}
    D_\text{KL}^c
    &= \frac{1}{\sum_{r=1}^{R} e^{\tau^c}}\sum^R_{r=1} e^{\tau_r^c}\Big[\frac{\sigma^2_{r,c}}{\sigma^2_{g,c}} - \ln{\frac{\sigma_{r,c}^2}{\sigma_{g,c}^2}}\Big]. 
\end{align}
We then find a close-form of $D_\text{KL}^c$ as follows:
\begin{align}
    D_\text{KL}^{c*} &= 
    \frac{1}{\sum_{r=1}^{R} e^{\tau^c}}\Big[\frac{\sum^R_{r=1} e^{\tau_r^c}\sigma^2_{r,c}}{\sigma^2_{g,c}} - \ln{\prod^R_{r=1}\Big(\frac{\sigma_{r,c}^2}{\sigma_{g,c}^2}}\Big)^{e^{\tau_r^c}}\Big] \notag \\
    &\overset{(a)}{\geq} 
    \frac{1}{\sum_{r=1}^{R} e^{\tau^c}}\Big[\frac{\sum^R_{r=1} e^{\tau_r^c}\sigma^2_{r,c}}{\sigma^2_{g,c}} - 
    \ln{\Big(\frac{\sum^R_{r=1} e^{\tau_r^c}\sigma^2_{r,c}}{\sigma^2_{g,c}}}\Big)\Big].
\label{eq:gaussian-LKD-label-1}
\end{align}
The inequality $(a)$ holds due to the Young's inequality for products \cite{2011-MF-YoungTheorem}. To understand the performance of LKD, we consider the optimal state of label-driven KL divergence (i.e., $D_\text{KL}^{c*} = \min{D_\text{KL}^c}$). Set $(\sum^R_{r=1} e^{\tau_r^c}\sigma^2_{r,c})/(\sigma^{*}_{g,c})^2 = u_r$. Then, we consider the function $f(x) = x - \ln{x}$. The optimal state of $f(x)$ is when derivative of $f(x)$ is $0$, which also means: $\nabla f(x) = 0$. Therefore, we have:
\begin{align}
    \nabla f(x) = 1 - \frac{1}{x} = 0
\label{eq:KL-optimal-state-variance}
\end{align}
Therefore, the function $f(x)$ receives the optimal state when $x = 1$. 

Taking the optimal result of \eqref{eq:KL-optimal-state-mean} and \eqref{eq:KL-optimal-state-variance}, we have the followings:
\begin{align}
    \sigma^{*2}_{\textrm{LKD},g,c} 
    = 
    \frac{1}{\sum_{r=1}^{R} e^{\tau^c}} \sum^R_{r=1} e^{\tau_r^c}\sigma^2_{r,c},
\label{eq:LKD-optimal-state-variance}
\end{align}
\begin{align}
    \mu^{*}_{\textrm{LKD},g,c} 
    = 
    \frac{1}{\sum_{r=1}^{R} e^{\tau^c}} \sum^R_{r=1} e^{\tau_r^c}\mu_{r,c}.
\label{eq:LKD-optimal-state-mean}
\end{align}

\section{Proof on Lemma~\ref{lemma:relationship-accuracy-variance}}
\label{appendix:relationship-accuracy-variance}
Let $b_c$ is the global optimal boundary to classify label $c$. We have the condition that the model $r$ have the accurate prediction on data that have label $c$ is $\mathcal{F}(x_i) \leq b_c$. Applying Central Limit Theorem (CLT) \cite{2010-MF-Probability}, we have: 
\begin{align}
    \tau^c_r = \rm{Pr}\Big[ \mathcal{F}(x_i) \leq b_c \Big] \geq 1 - \frac{1}{\sqrt{2\pi}} \exp{\Big[-\frac{1}{2}\left(\frac{b_c}{\sigma_{r,c}}\right)^2\Big]}. 
\label{eq:accuracy-to-variance}
\end{align}
As we can see from \eqref{eq:accuracy-to-variance}, $\tau^c_r$ is proportional to $1/\sigma^2_{r,c}$. Thus we have when $\sigma^2_{1,c} \leq \sigma^2_{2,c} \leq \dots \leq \sigma^2_{R,c}$, the model accuracy must satisfy the constraint $\tau^2_{1,c} \geq \tau^2_{2,c} \geq \dots \geq \tau^2_{R,c}$.

\section{Proof on Lemma~\ref{lemma:regional-DPGMM}}
\label{appendix:regional-DPGMM}
Due to the non-IID settings on FL, the data class can be distributed as Dirichlet process \cite{2019-FL-BayesianNonParametric}. To be more detailed, the data distribution can be represented as follows: 
\begin{align}
    p\left(x \vert y = c \right) = \pi_{j}.
\label{eq:dirichlet-data-distribution}
\end{align}
We have $\pi_j$ is the Dirichlet allocation weight of data on ground truth with label $l$. 
\begin{align}
    \pi^{l}_{j} = \frac{1}{\mathcal{B}(\nu)} \prod^{C}_{c=1} x_c^{\nu_c-1} \quad \text{where} \quad \mathcal{B}(\nu)= \frac{\prod^{C}_{c}\Gamma(\nu)}{\Gamma \Big(\sum^{C}_{c=1} \nu_c \Big)},
\end{align}
where $\Gamma(\nu)$ is the gamma function and $\nu$ is the Dirichlet coefficient. Applying \eqref{eq:dirichlet-data-distribution} into Lemma~\ref{assumption:data-DPGMM}, we have the following equation:
\begin{align}
    p\left(x \vert (\mu_c, \Sigma_c, \pi_c)^C_{c=1}\right) 
    &= 
    \sum^{C}_{c=1} \pi_{c} \sum^{S_c^r}_{j=1} ~ \mathcal{N}(\mathbf{x}; \mu_c, \Sigma_c) \\
    &= 
    \sum^{C}_{c=1} \pi_{c} \pi_{l} S_c^r ~ \mathcal{N}(\mathbf{x}; \mu_c, \Sigma_c) \\
    &=
    \sum^{C}_{c=1} \alpha^r_c ~ \mathcal{N}(\mathbf{x}; \mu_c, \Sigma_c).
\label{eq:dirichlet-data-on-client}
\end{align}
Here, we denote $\alpha^r_c = \pi_{c} \pi_{l} S_c^r$. Since $\pi_{c}, \pi_{l}$ follow Dirichlet distribution, and $S_c^r$ is a constant, $\pi_{c,j}$ follows Dirichlet distribution.

\clearpage
\section{Data and Parameter Settings}
\label{appendix:data-parameter-settings}
\begin{table}[!h]
\renewcommand{\arraystretch}{1.25}
\caption{The data settings used for evaluation.\\}
\centering 
\small\addtolength{\tabcolsep}{-3pt}
\begin{tabular}{| l | c | c | c | c | c | c |} 
\hline
\multicolumn{7}{c}{\textbf{Data Setting}} \\
\hline
& MNIST & EMNIST  & CIFAR-10 & CIFAR-100 & CINIC-10 & CelebA\\ [0.5ex] 
\hline  
Data size          & $48000$  & $118440$ & $48000$ & $48000$  & $48000$ & $200000$ \\ 
Regions            & $3$ & $3$ & $3$ & $3$ & $3$ & $3$       \\ 
non-IID coeff $\alpha$ & $0.1,1$ & $0.1,1$ & $0.1,1$ & $0.1,1$ & $0.1,1$ & $0.1,1$        \\  
\hline
\multicolumn{7}{c}{\textbf{Federated Learning system}} \\
\hline
Clients per region  & $10$ & $10$ & $10$ & $10$ & $10$ & $10$        \\  
Samples per client  & $1600$  & $3948$ & $1600$ & $1600$ & $1600$ & $5000$        \\  
Samples on server  & $3200$  & $8000$ & $3200$ & $3200$  & $3200$ & $10000$       \\  
\hline
\multicolumn{7}{c}{\textbf{Client settings}} \\
\hline
Training models & LeNet-$5$ & LeNet-$5$ & ResNet-$50$ & ResNet-$50$  & ResNet-$50$ & ResNet-$50$ \\  
\hline
Epochs         & $5$ & $5$ & $5$ & $5$ & $5$ & $5$        \\  
Learning rate  & $0.05$ & $0.05$ & $0.05$ & $0.05$ & $0.05$ & $0.05$       \\  
\hline 
\end{tabular}
\label{tab:data-configuration} 
\end{table}
\clearpage
\section{Experimental Results on Soft-loss Contribution}
\begin{table*}[!h]
\renewcommand{\arraystretch}{1.25}
\caption{Top-1 accuracy of F2L on dataset EMNIST with different hard-loss scaling coefficients $\lambda_3$. The soft-loss $\lambda_1$ and $\lambda_2$ is scaled to be $1-\lambda_3$ for a clear evaluation. The teacher efficiency is observed at the $5^\text{th}$ distillation round (which is at every 40 rounds of communication).\\} 
\centering 
\begin{tabular}{|l|c|c|c|c|c|c|}
\hline
&
  $\lambda_3 = 0$
& $\lambda_3 = 0.001$ 
& $\lambda_3 = 0.01$
& $\lambda_3 = 0.1$
& $\lambda_3 = 0.5$
& $\lambda_3 = 1$\\ \cline{1-7} 

Teacher 1      & $64.33$ & $64.33$ & $64.33$ & $64.33$ & $64.33$ & $64.33$ \\ \hline
Teacher 2      & $73.09$ & $73.09$ & $73.09$ & $73.09$ & $73.09$ & $73.09$ \\ \hline
Teacher 3      & $74.66$ & $74.66$ & $74.66$ & $74.66$ & $74.66$ & $74.66$ \\ \hline
G-student      & $71.75$ & $82.96$ & $\mathbf{85.87}$ & $\mathbf{85.20}$ & $84.30$ & $82.29$   \\ \hline
\end{tabular}
\label{tab:LKD-DistillationTest-EMNIST} 
\end{table*}
\begin{table*}[!h]
\renewcommand{\arraystretch}{1.25}
\caption{Top-1 accuracy of F2L on dataset CIFAR-10 with different hard-loss scaling coefficients $\lambda_3$. The soft-loss $\lambda_1$ and $\lambda_2$ is scaled to be $1-\lambda_3$ for a clear evaluation. The teacher efficiency is observed at the $5^\text{th}$ distillation round (which is at every 40 rounds of communication).\\} 
\centering 
\begin{tabular}{|l|c|c|c|c|c|c|}
\hline
&
  $\lambda_3 = 0$
& $\lambda_3 = 0.001$ 
& $\lambda_3 = 0.01$
& $\lambda_3 = 0.1$
& $\lambda_3 = 0.5$
& $\lambda_3 = 1$\\ \cline{1-7} 

Teacher 1      & $32.11$  & $32.11$ & $32.11$ & $32.11$ & $32.11$ & $32.11$\\ \hline
Teacher 2      & $30.64$  & $30.64$ & $30.64$ & $30.64$ & $30.64$ & $30.64$\\ \hline
Teacher 3      & $27.53$  & $27.53$ & $27.53$ & $27.53$ & $27.53$ & $27.53$\\ \hline
G-student      & $25.26$  & $\mathbf{41.32}$ & $\mathbf{52.71}$ & $49.98$ & $48.80$ & $47.98$\\ \hline
\end{tabular}
\label{tab:LKD-DistillationTest-CIFAR10} 
\end{table*}
\begin{table*}[!h]
\renewcommand{\arraystretch}{1.25}
\caption{Top-1 accuracy of F2L on dataset CIFAR-100 with different hard-loss scaling coefficients $\lambda_3$. The soft-loss $\lambda_1$ and $\lambda_2$ is scaled to be $1-\lambda_3$ for a clear evaluation. The teacher efficiency is observed at the $5^\text{th}$ distillation round (which is at every 40 rounds of communication).\\} 
\centering 
\begin{tabular}{|l|c|c|c|c|c|c|}
\hline
&
  $\lambda_3 = 0$
& $\lambda_3 = 0.001$ 
& $\lambda_3 = 0.01$
& $\lambda_3 = 0.1$
& $\lambda_3 = 0.5$
& $\lambda_3 = 1$\\ \cline{1-7} 

Teacher 1      & $6.62$ & $6.62$ & $6.62$ & $6.62$ & $6.62$ & $6.62$  \\ \hline
Teacher 2      & $6.15$ & $6.15$ & $6.15$ & $6.15$ & $6.15$ & $6.15$  \\ \hline
Teacher 3      & $7.56$ & $7.56$ & $7.56$ & $7.56$ & $7.56$ & $7.56$  \\ \hline
G-student      & $7.40$ & $\mathbf{15.08}$ & $\mathbf{15.26}$ & $\mathbf{14.33}$ & $11.44$ & $9.64$   \\ \hline
\end{tabular}
\label{tab:LKD-DistillationTest-CIFAR100} 
\end{table*}

\clearpage
\section{Experimental Results on Required Training Sample Size}
\begin{table*}[!h]
\renewcommand{\arraystretch}{1.25}
\caption{Top-1 accuracy of F2L on dataset EMNIST with different data-on-server numbers of samples. $\delta$ represents the sample scaling ratio compared to the data-on-server as demonstrated in Table~\ref{tab:data-configuration}. The teacher efficiency is observed at the $5^\text{th}$ distillation round (which is at every 40 rounds of communication).\\} 
\centering 
\begin{tabular}{|l|c|c|c|c|c|c|}
\hline
&
  $\delta = 1$
& $\delta = 1/2$ 
& $\delta = 1/4$
& $\delta = 1/6$
& $\delta = 1/8$
& $\delta = 1/10$\\ \cline{1-7} 

Teacher 1      & $64.33$ & $64.33$ & $64.33$ & $64.33$ & $64.33$ & $64.33$ \\ \hline
Teacher 2      & $73.09$ & $73.09$ & $73.09$ & $73.09$ & $73.09$ & $73.09$ \\ \hline
Teacher 3      & $74.66$ & $74.66$ & $74.66$ & $74.66$ & $74.66$ & $74.66$ \\ \hline
G-student      & $\mathbf{85.20}$ & $\mathbf{83.96}$ & $80.71$ & $79.48$ & $76.64$ & $74.22$   \\ \hline
\end{tabular}
\label{tab:LKD-SampleTrainTest-EMNIST} 
\end{table*}
\begin{table*}[!h]
\renewcommand{\arraystretch}{1.25}
\caption{Top-1 accuracy of F2L on dataset CIFAR-$10$ with different data-on-server numbers of samples. $\delta$ represents the sample scaling ratio compared to the data-on-server as demonstrated in Table~\ref{tab:data-configuration}. The teacher efficiency is observed at the $5^\text{th}$ distillation round (which is at every 40 rounds of communication).\\} 
\centering 
\begin{tabular}{|l|c|c|c|c|c|c|}
\hline
&
  $\delta = 1$
& $\delta = 1/2$ 
& $\delta = 1/4$
& $\delta = 1/6$
& $\delta = 1/8$
& $\delta = 1/10$\\ \cline{1-7} 

Teacher 1      & $32.11$  & $32.11$ & $32.11$ & $32.11$ & $32.11$ & $32.11$\\ \hline
Teacher 2      & $30.64$  & $30.64$ & $30.64$ & $30.64$ & $30.64$ & $30.64$\\ \hline
Teacher 3      & $27.53$  & $27.53$ & $27.53$ & $27.53$ & $27.53$ & $27.53$\\ \hline
G-student      & $\mathbf{54.71}$  & $\mathbf{52.04}$ & $42.22$ & $38.18$ & $37.48$ & $34.54$\\ \hline
\end{tabular}
\label{tab:LKD-SampleTrainTest-CIFAR10} 
\end{table*}
\begin{table*}[!h]
\renewcommand{\arraystretch}{1.25}
\caption{Top-1 accuracy of F2L on dataset CIFAR-$100$ with different data-on-server numbers of samples. $\delta$ represents the sample scaling ratio compared to the data-on-server as demonstrated in Table~\ref{tab:data-configuration}. The teacher efficiency is observed at the $5^\text{th}$ distillation round (which is at every 40 rounds of communication).\\} 
\centering 
\begin{tabular}{|l|c|c|c|c|c|c|}
\hline
&
  $\delta = 1$
& $\delta = 1/2$ 
& $\delta = 1/4$
& $\delta = 1/6$
& $\delta = 1/8$
& $\delta = 1/10$\\ \cline{1-7} 

Teacher 1      & $6.62$ & $6.62$ & $6.62$ & $6.62$ & $6.62$ & $6.62$ \\ \hline
Teacher 2      & $6.15$ & $6.15$ & $6.15$ & $6.15$ & $6.15$ & $6.15$ \\ \hline
Teacher 3      & $7.56$ & $7.56$ & $7.56$ & $7.56$ & $7.56$ & $7.56$ \\ \hline
G-student      & $\mathbf{15.41}$ & $14.98$ & $14.33$ & $11.25$ & $10.83$ & $9.51$   \\ \hline
\end{tabular}
\label{tab:LKD-SampleTrainTest-CIFAR100} 
\end{table*}

\end{document}